\documentclass{article}

\usepackage{iclr2027_conference,times}

\usepackage{amsmath,amsfonts,bm}

\def\eqref#1{equation~\ref{#1}}
\def\1{\bm{1}}

\DeclareMathAlphabet{\mathsfit}{\encodingdefault}{\sfdefault}{m}{sl}
\SetMathAlphabet{\mathsfit}{bold}{\encodingdefault}{\sfdefault}{bx}{n}

\renewcommand{\eqref}[1]{(\ref{#1})}

\usepackage[utf8]{inputenc}
\usepackage[T1]{fontenc}
\usepackage{microtype}

\usepackage{hyperref}
\hypersetup{hidelinks}
\usepackage{url}

\usepackage{graphicx}
\usepackage{booktabs}
\usepackage{multirow}
\usepackage{tabularx}
\usepackage{tabulary}
\usepackage{multicol}
\usepackage{array}

\usepackage{xcolor}
\usepackage{colortbl}

\usepackage{bbm}
\usepackage{subcaption}

\usepackage{amsmath}
\usepackage{amssymb}
\usepackage{amsfonts}
\usepackage{amsthm}
\usepackage{bm}

\usepackage{algorithm}
\usepackage{float}
\usepackage[noend]{algpseudocode}

\usepackage{pifont}

\usepackage{tikz}
\usepackage{wrapfig}
\usetikzlibrary{
    arrows.meta,
    positioning,
    calc,
    fit,
    backgrounds
}

\definecolor{lavender}{RGB}{230,230,250}
\definecolor{softgray}{RGB}{245,245,245}
\definecolor{softgreen}{RGB}{228,242,231}

\newcommand{\cmark}{\ding{51}}
\newcommand{\xmark}{\ding{55}}

\renewcommand{\cmark}{\textcolor{green!60!black}{\ding{51}}}
\renewcommand{\xmark}{\textcolor{red!70!black}{\ding{55}}}

\title{SlackDrive: Reclaiming Runtime Slack for Adaptive Driving Inference}

\author{%
  \makebox[\textwidth][c]{%
    \begin{tabular}{c}
      Xiaohuan Pei\textsuperscript{1}\quad
      Hengguang Zhou\textsuperscript{2}\quad
      Yuanhao Ban\textsuperscript{2}\quad
      Justin Cui\textsuperscript{2}\quad
      Jiaqi Feng\textsuperscript{2}
      \\[2pt]
      Haoyu Xie\textsuperscript{2}\quad
      Tao Huang\textsuperscript{3}\quad
      Pichao Wang\textsuperscript{4}\quad
      Yanchao Yang\textsuperscript{5}\quad
      Cho-Jui Hsieh\textsuperscript{2}
      \\[6pt]
      {\normalfont\small
      \textsuperscript{1}The University of Sydney
      \qquad
      \textsuperscript{2}University of California, Los Angeles}
      \\[2pt]
      {\normalfont\small
      \textsuperscript{3}Shanghai Jiao Tong University
      \qquad
      \textsuperscript{4}NVIDIA
      \qquad
      \textsuperscript{5}The University of Hong Kong}
    \end{tabular}%
  }
}

\iclrfinalcopy

\begin{document}

\maketitle

% Remove:
% "Published as a conference paper at ICLR 2027"
\lhead{}

% ============================================================
% Main paper
% ============================================================

\begin{abstract}
Driving world-action models improve planning by coupling multimodal reasoning with future prediction, but their growing inference cost increasingly conflicts with the real-time latency requirements of vehicle control.
Existing acceleration methods reduce tokens, layers, or sampling steps with policies selected prior to deployment, yet leave residual runtime variation largely unexploited after offline profiling and static scheduling on shared onboard compute.
We observe that the largest admissible compute budget varies systematically with the residual runtime state, while recent realized latency provides a direct signal of the available compute slack.
Motivated by this observation, we propose \textbf{SlackDrive}, a pre-inference compute allocator that reuses realized latency to select the compute budget of each control step before model execution.
SlackDrive profiles the latency and planning utility of a small discrete budget set once, estimates online compute state from completed forwards, and selects the highest-utility budget predicted to remain within the admissible latency envelope, complementing existing profiling and resource scheduling while preserving the driving backbone and its compute actuator.
On NAVSIM v2 with DriveDreamer-Policy, SlackDrive improves latency-constrained EPDMS by $21.7\%$ over the strongest baseline under a stringent latency regime, while the full-budget model and preconfigured token-pruning baselines exceed the admissible latency envelope under runtime contention.
\end{abstract}

\section{Introduction}
\label{sec:Introduction}

End-to-end autonomous driving has progressed from vision-centric planners~\cite{chitta2022transfuser,hu2023uniad,weng2024paradrive,liao2025diffusiondrive,jia2025drivetransformer} to vision-language and world-action policies~\cite{tian2024drivevlm,yang2025drivemoe,li2025recogdrive,zhou2025autovla,zhou2026drivedreamerpolicy,liu2026uniworldvla,wang2026latentwam,shi2026drivewam} that combine semantic reasoning, future modeling, and trajectory generation in a unified policy.
In a mainstream driving foundation-model pipeline, multi-view images and navigation context are encoded into multimodal representations, a policy or world module reasons over the current scene and possible futures, and an action head decodes these representations into an ego trajectory.
However, dense visual context, large reasoning backbones, and iterative generative heads introduce substantial per-step latency, so a compute setting chosen for an unloaded accelerator can become infeasible once the same model is placed inside a latency-constrained control loop.

\begin{figure*}[t]
    \centering

    % ===================== Main content =====================
    \begin{minipage}[t]{0.655\textwidth}
        \vspace{0pt}
        \centering

        \includegraphics[
            width=\linewidth
        ]{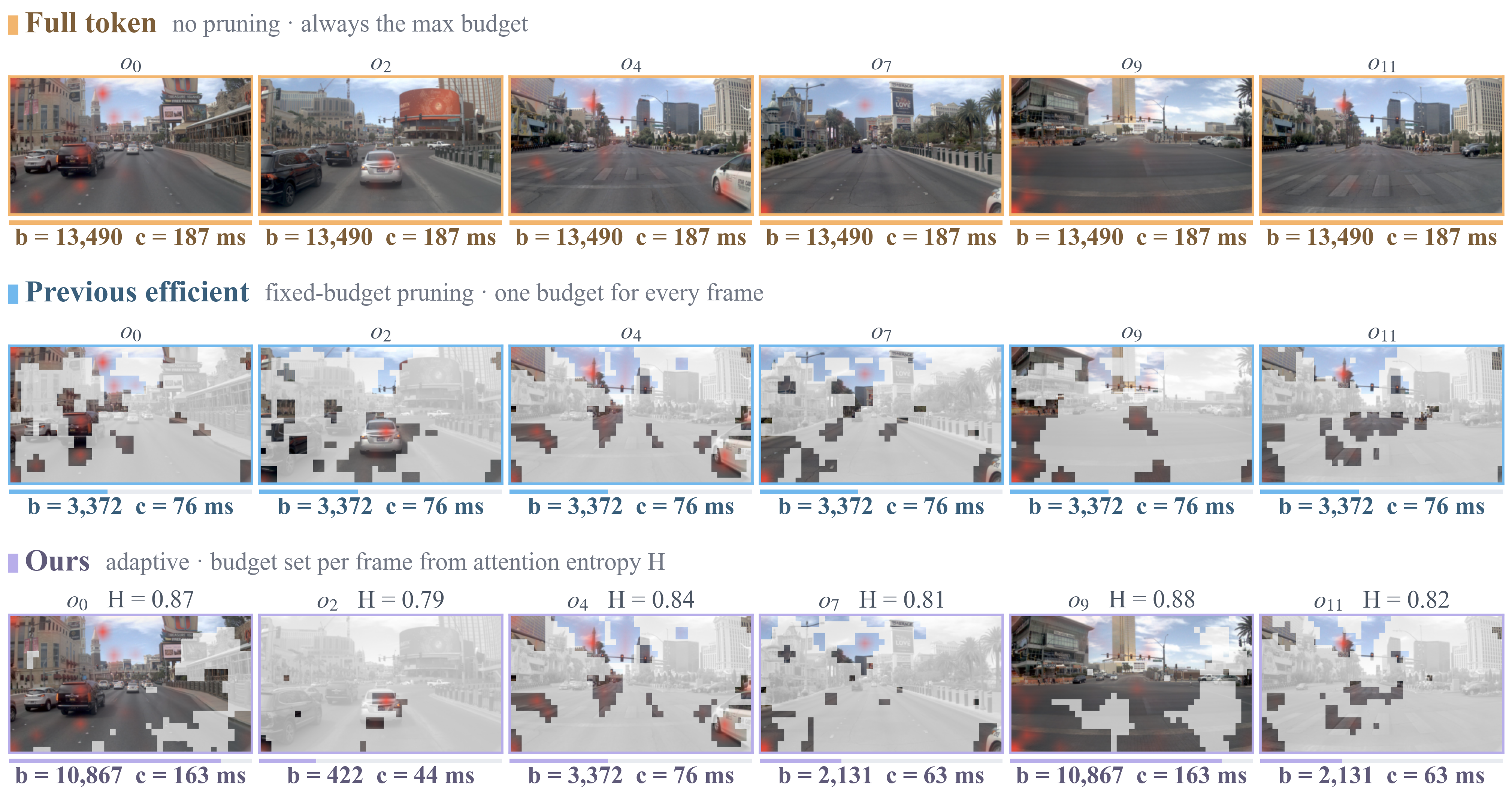}
    \end{minipage}
    \hfill
    \begin{minipage}[t]{0.325\textwidth}
        \vspace{2mm}
        \centering

        % ---------- (b) ----------
        \includegraphics[
            width=\linewidth
        ]{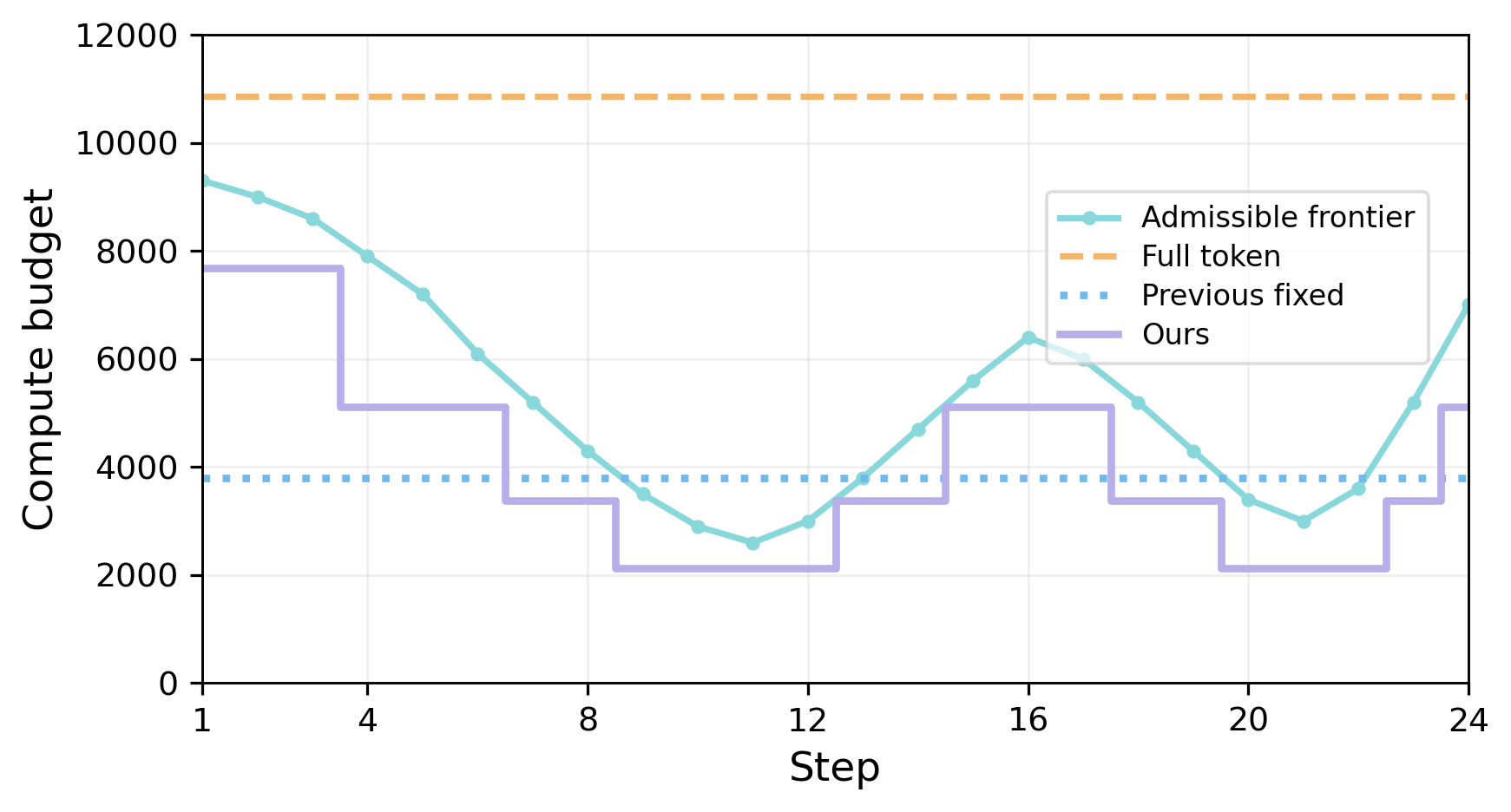}

        \vspace{-1mm}
        {\small\bfseries (b) Online compute allocation}

        \vspace{1.5mm}

        % ---------- (c) table ----------
        {
        \renewcommand{\arraystretch}{1.58}
        \resizebox{\linewidth}{!}{
            \begin{tabular}{l l}
                \toprule
                \textbf{Model} & \textbf{Conditional generation} \\
                \midrule

                Vision-Language-Action
                & $\hat{\mathbf A}_t
                \sim f_\theta^A(\cdot \mid o_t,g_t)$ \\

                World Model
                & $\hat{s}_{t+1}
                \sim f_\theta^W(\cdot \mid s_t,\mathbf A_t)$ \\

                World-Action Model
                & $(\hat{s}_{t+1},\hat{\mathbf A}_t)
                \sim f_\theta^{WA}(\cdot \mid s_t,o_t,g_t)$ \\

                \rowcolor{lavender}
                \textbf{Ours}
                & $(\hat{s}_{t+1},\hat{\mathbf A}_t)
                \sim \widetilde f_{\theta,\pi}
                (\cdot \mid s_t,o_t,g_t,\mathbf c_{t-1})$ \\

                \bottomrule
            \end{tabular}
        }
        }
    \end{minipage}

    % ===================== Aligned panel labels =====================
    \vspace{0.8mm}

    \begin{minipage}[t]{0.655\textwidth}
        \centering
        {\small\bfseries (a) Full vs. Static vs. Dynamic}
    \end{minipage}
    \hfill
    \begin{minipage}[t]{0.325\textwidth}
        \centering
        {\small\bfseries (c) Previous vs. Ours}
    \end{minipage}

    \vspace{1.2mm}

    \caption{
    \textbf{Motivation and formulation.}
    \textbf{(a)} Full-token uses maximum compute, prior efficient methods use a fixed budget, and SlackDrive adapts the budget per step.
    \textbf{(b)} The admissible budget changes with measured inference latency. SlackDrive selects the largest feasible budget before the next forward pass.
    \textbf{(c)} SlackDrive conditions world-action inference on the measured latency state $\mathbf c_{t-1}$ with a frozen backbone.
    }
    \vspace{-3mm}
    \label{fig:motivation}
\end{figure*}

Existing work pursues faster driving inference through architectural parallelism and selective computation, such as FastDriveVLA~\cite{cao2025fastdrivevla} that focuses on reconstruction-guided visual-token pruning, Prune2Drive~\cite{xiong2025prune2drive} that aims at multi-view token reduction, ST-Prune~\cite{sha2026stprune} that targets spatio-temporal redundancy, DiffusionDrive~\cite{liao2025diffusiondrive} that reduces iterative trajectory generation, and AutoMoT~\cite{huang2026automot} that uses asynchronous execution frequencies across reasoning and action branches.
% However, a key but underexplored property of real-time driving is that
% \textbf{inference latency is usually treated as an outcome, while its realized value can be reused as a runtime state to condition the next compute allocation}.
% As Figure~\ref{fig:motivation} shows, residual runtime variation can change the largest admissible compute budget even after the model and its operating points have been profiled offline.
% This signal is available from every completed inference.
% Existing efficiency methods ignore this measured latency and therefore commit to compute without knowing how much residual runtime headroom is currently available.
However, a key but underexplored property of real-time driving is that
\textbf{inference latency is usually treated as an outcome, while its realized value can be reused as a runtime state to condition the next compute allocation}.
As Figure~\ref{fig:motivation} shows, residual runtime variation can change the largest admissible compute budget even after the model and its operating points have been profiled offline, as also observed in recent multi-DNN edge inference systems under runtime contention~\cite{han2024pantheon,ling2025time,han2024latency}.
Realized latency is available after every completed forward, yet existing efficiency methods do not use it to infer the compute headroom available for the next inference.

To address this challenge, we introduce \textbf{SlackDrive}, a plug-and-play pre-inference compute allocator that adapts an existing inference knob to residual runtime variation while keeping the driving model frozen.
SlackDrive is built on two complementary ideas.
First, \emph{budget profiling} measures, once, the latency and planning utility of a small set of admissible compute budgets and retains their non-dominated quality and cost operating points.
Second, \emph{online compute-state estimation} converts completed-forward latency into a normalized runtime state and propagates its local level and uncertainty to the next control step.
When recent inference is fast and stable, the admissible set expands and SlackDrive restores a larger budget to preserve planning quality.
Conversely, when runtime load increases, the admissible set contracts and SlackDrive lowers the budget before inference, rather than observing a latency violation only after an expensive forward pass has completed.
The resulting controller requires only scalar state updates and a scan over a small budget table, making it orthogonal to whether the underlying actuator changes visual tokens, reasoning depth, action queries, or diffusion steps.

Our contributions can be summarized as follows.
(1) We identify residual runtime-dependent compute admissibility as a distinct source of inefficiency in driving foundation models and show that the computation supported by a fixed control period varies with the runtime state.
This observation motivates online compute allocation that complements offline profiling and existing acceleration methods, which primarily optimize what computation to remove rather than how much computation the current step can execute.
(2) We propose \textbf{SlackDrive}, which combines one-time quality and latency profiling with online compute-state estimation to select the highest-utility latency-feasible budget before each model invocation.
(3) We formulate latency-constrained EPDMS and evaluate SlackDrive on NAVSIM v2 with DriveDreamer-Policy, where it improves EPDMS@$130$\,ms from $63.03$ to $76.71$, a $21.7\%$ relative gain over the strongest baseline under the same latency constraint.

\section{Related Work}
\label{sec:related}

\textbf{Driving world-action models.}
End-to-end driving models increasingly combine perception, semantic reasoning, future prediction, and trajectory planning inside a shared learned policy.
Early planning-oriented systems such as TransFuser~\cite{chitta2022transfuser}, UniAD~\cite{hu2023uniad}, and PARA-Drive~\cite{weng2024paradrive} learn trajectories directly from sensor representations, while DiffusionDrive~\cite{liao2025diffusiondrive} introduces truncated diffusion for multimodal trajectory generation.
Vision-language driving models extend this pipeline with pretrained semantic knowledge, including DriveVLM~\cite{tian2024drivevlm}, DriveMoE~\cite{yang2025drivemoe}, ReCogDrive~\cite{li2025recogdrive}, and AutoVLA~\cite{zhou2025autovla}.
A parallel line introduces future prediction as an explicit planning variable through world-action models such as DriveDreamer-Policy~\cite{zhou2026drivedreamerpolicy}, Uni-World VLA~\cite{liu2026uniworldvla}, Latent-WAM~\cite{wang2026latentwam}, DriveWAM~\cite{shi2026drivewam}, and GigaWorld-Policy~\cite{ye2026gigaworldpolicy}.
These models expose several natural compute knobs, including the number of visual or query tokens, the frequency of semantic reasoning, and the number of iterative generative steps.
EfficientDrive does not modify how these models learn world or action representations; it treats such a knob as an actuator and allocates its value from the current compute state before each forward pass.

\textbf{Efficient driving inference.}
The inference cost of multimodal driving and perception has motivated efficient architectures~\cite{liu2025slam3r,dong2025reloc3r} and selective computation.
FastDriveVLA~\cite{cao2025fastdrivevla}, Prune2Drive~\cite{xiong2025prune2drive}, VLA-ADP~\cite{pei2026action}, and ST-Prune~\cite{sha2026stprune} reduce visual-token redundancy, while CSP~\cite{pei2024cross} and causal-mask attention~\cite{pei2026rethinking} reduce cache and attention overhead.
DiffusionDrive~\cite{liao2025diffusiondrive} shortens iterative trajectory generation, while AutoMoT~\cite{huang2026automot} and MindVLA-U1~\cite{huang2026mindvlau1} use asynchronous or fast/slow execution paths to reduce unnecessary computation.
BLUE~\cite{ling2026blue} gates language generation on a per-frame basis.
These approaches answer which model operations are unnecessary for a given architecture or scene, and their policies can be directly used as the candidate budget configurations in our framework.
However, they do not explicitly close the loop on measured device latency, leaving the compute level unchanged when identical model computation becomes slower or faster because runtime conditions change.

\subsection{Preliminaries}
\label{subsec:preliminary}

\textbf{Driving inference.}
We consider three common formulations for learning-based driving inference.
Given the current observation $o_t$, navigation goal $g_t$, world state $s_t$, and action sequence $\mathbf A_t$, a vision-language-action model directly predicts future actions, a world model predicts the next state conditioned on actions, and a world-action model jointly predicts future state and actions:
\begin{equation}
\label{eq:driving_formulation}
\begin{aligned}
\text{VLA:}\qquad
&\hat{\mathbf A}_{t}
\sim
f^{\mathrm A}_{\theta}
\!\left(\cdot\mid o_t,g_t\right),\\
\text{World Model:}\qquad
&\hat s_{t+1}
\sim
f^{\mathrm W}_{\theta}
\!\left(\cdot\mid s_t,\mathbf A_t\right),\\
\text{World-Action Model:}\qquad
&(\hat s_{t+1},\hat{\mathbf A}_{t})
\sim
f^{\mathrm{WA}}_{\theta}
\!\left(\cdot\mid s_t,o_t,g_t\right).
\end{aligned}
\end{equation}
Here $\theta$ denotes the model parameters.
We use the world-action formulation as the running example because it combines multimodal reasoning, future prediction, and action generation within a single inference process.

\textbf{Latency-constrained inference.}
Let $\mathcal B$ denote a finite set of inference configurations and $b_t\in\mathcal B$ the compute budget selected at step $t$.
Conditioning the frozen world-action model on $b_t$ changes the amount of computation used by the forward pass without changing its semantic inputs:
\begin{equation}
\label{eq:budgeted_inference}
\begin{aligned}
(\hat s_{t+1}^{(b_t)},\hat{\mathbf A}_{t}^{(b_t)})
&\sim
f^{\mathrm{WA}}_{\theta}
\!\left(
\cdot\mid s_t,o_t,g_t;b_t
\right),\\
\ell_t(b_t)
&=
L(b_t,\omega_t),
\end{aligned}
\end{equation}
where $\ell_t(b_t)$ is the realized inference latency and $\omega_t$ denotes the runtime condition of the accelerator.
The semicolon distinguishes the inference configuration $b_t$ from the semantic conditioning variables of the driving model.

Let $U(b)$ denote the planning utility associated with configuration $b$, $\tau_{\max}$ the admissible per-step latency, and $\delta$ the target violation probability.
The latency-constrained allocation problem is
\begin{equation}
\label{eq:chance_objective}
\begin{aligned}
b_t^{\star}
\in
\arg\max_{b\in\mathcal B}
\quad & U(b)\\
\text{s.t.}\quad
&
\Pr\!\left[
L(b,\omega_t)\le\tau_{\max}
\mid
\mathcal H_{t-1}^{\mathrm{run}}
\right]
\ge 1-\delta,
\end{aligned}
\end{equation}
where $\mathcal H_{t-1}^{\mathrm{run}}$ contains the completed runtime observations available before the current forward.
Eq.~\ref{eq:chance_objective} therefore asks for the highest-utility compute configuration that remains admissible under the current latency constraint.

To evaluate planning quality jointly with real-time feasibility, we define latency-constrained EPDMS as
\begin{equation}
\label{eq:epdms_latency}
\mathrm{EPDMS}@\tau_{\max}
=
\frac{1}{N}
\sum_{i=1}^{N}
q_i(b_i)\,
\mathbbm{1}
\!\left[
\ell_i(b_i)\le\tau_{\max}
\right],
\end{equation}
where $q_i(b_i)$ is the EPDMS contribution of scene $i$ under budget $b_i$, $N$ is the number of evaluated scenes, and $\mathbbm{1}[\cdot]$ is the indicator function.
A prediction contributes to the score only when it completes within the admissible latency envelope.

\section{Methodology}
\label{sec:method}

\begin{figure*}[t]
    \centering
    \includegraphics[width=\linewidth]{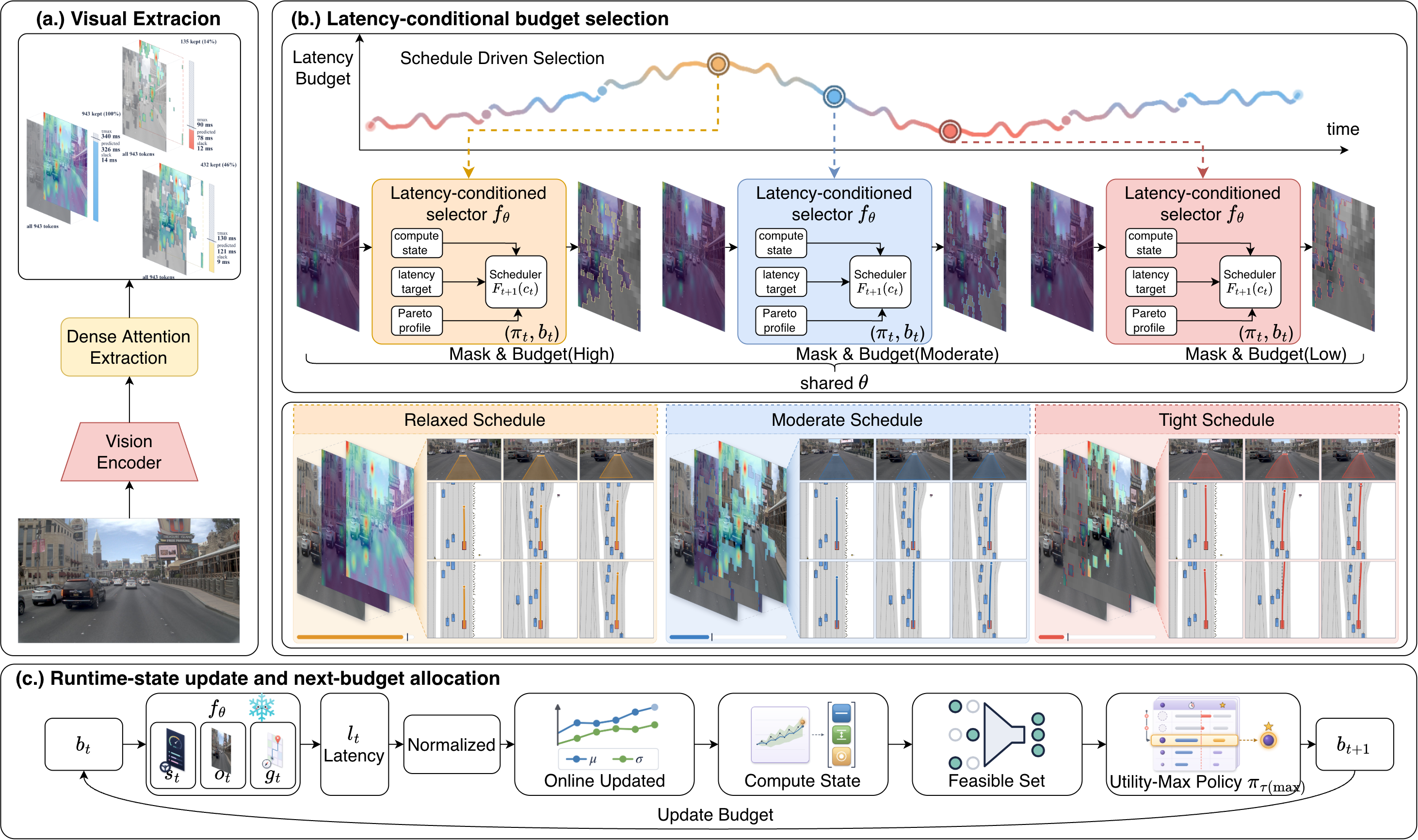}
    % \fbox{
    %     \parbox[c][0.185\textheight][c]{0.965\textwidth}{
    %         \centering
    %         {\Large\bfseries Method Overview Placeholder}\\[3mm]
    %         {\normalsize
    %         Runtime history
    %         $\rightarrow$
    %         Compute-state modeling
    %         $\rightarrow$
    %         Latency-feasible set
    %         $\rightarrow$
    %         Pre-inference allocation
    %         $\rightarrow$
    %         Frozen world-action model
    %         $\rightarrow$
    %         Latency feedback}
    %     }
    % }
    \vspace{-5mm}
    \caption{\textbf{Overview of SlackDrive.}
     \textbf{(a) Visual extraction} obtains dense attention cues from the vision encoder for subsequent token selection.
    (\textbf{b) Latency-conditional budget selection} adapts the attention mask and compute budget to the current latency target, yielding different configurations under relaxed, moderate, and tight schedules.
    \textbf{(c) Runtime-state update and next-budget allocation} feeds the realized inference latency back into the runtime state, from which a utility-maximizing policy selects the feasible budget for the next step.}
    \vspace{-3mm}
    \label{fig:method_overview}
\end{figure*}

SlackDrive realizes Eq.~\ref{eq:chance_objective} via a scheduler: it profiles the discrete latency frontier once, converts completed runtime observations into a compact compute state, and uses that state to select the next inference configuration before the expensive forward begins.

\textbf{Compute-state modeling.}
The key implementation is to replace the unavailable accelerator condition $\omega_{t+1}$ with a compact state inferred from completed forwards.
Let
$\mathcal H^{\mathrm{run}}_t:=\{(b_j,\ell_j)\}_{j=1}^{t}$
denote the completed runtime history.
SlackDrive summarizes $\mathcal H^{\mathrm{run}}_t$ as $\mathbf c_t\in\mathbb R^3$, lets an external policy $\pi_{\tau_{\max}}$ select the next configuration from an offline profile $\mathcal P$, and composes this policy with the frozen world-action model:
\begin{equation}
\label{eq:composite_inference}
\begin{aligned}
\mathbf c_t
&=\Phi_\alpha\!\left(\mathcal H^{\mathrm{run}}_t\right),
\qquad
b_{t+1}=\pi_{\tau_{\max}}(\mathbf c_t;\mathcal P),\\[-1pt]
\widetilde f_{\theta,\pi}
\!\left(\,\cdot\mid s_{t+1},o_{t+1},g_{t+1},\mathbf c_t;\tau_{\max}\right)
&:=
f^{\mathrm{WA}}_{\theta}
\!\left(\,\cdot\mid s_{t+1},o_{t+1},g_{t+1};
\pi_{\tau_{\max}}(\mathbf c_t;\mathcal P)\right),\\[-1pt]
\big(\hat s_{t+2},\hat{\mathbf A}_{t+1}\big)
&\sim
\widetilde f_{\theta,\pi}
\!\left(\,\cdot\mid s_{t+1},o_{t+1},g_{t+1},\mathbf c_t;\tau_{\max}\right).
\end{aligned}
\end{equation}
Thus $\mathbf c_t$ conditions only the external compute policy, not $f^{\mathrm{WA}}_\theta$ itself.
The remainder of this section specifies $\mathcal P$, $\Phi_\alpha$, and $\pi_{\tau_{\max}}$.

\subsection{Budget Profile}
\label{subsec:profile}

\textbf{Quality and latency frontier.}
For every $b_k\in\mathcal B$, we execute the frozen model under a nominal device state and record its reference latency $L_0(b_k)\in\mathbb R_{>0}$ and validation utility $U(b_k)\in\mathbb R$.
Rather than assume that more raw computation is always more useful, we remove dominated operating points and retain the Pareto profile
\begin{equation}
\label{eq:pareto_profile}
\begin{aligned}
\mathcal B_{\mathrm F}
&=\Big\{b\in\mathcal B:\nexists b'\in\mathcal B\ \text{s.t.}\
L_0(b')\le L_0(b),\ U(b')\ge U(b),\\[-1pt]
&\hspace{42mm}
\big[L_0(b')<L_0(b)\ \vee\ U(b')>U(b)\big]\Big\},\\[-1pt]
\mathcal P
&=\Big\{\big(b,L_0(b),U(b)\big):b\in\mathcal B_{\mathrm F}\Big\}.
\end{aligned}
\end{equation}
The profile is model-specific but is collected only once and contains no learned parameters of SlackDrive.
Eq.~\ref{eq:pareto_profile} guarantees that the online policy never selects a configuration for which another profiled operating point is both no slower and no less accurate.

\subsection{Compute-State Allocation}
\label{subsec:scheduling}

\textbf{Normalized runtime load.}
Direct latency values cannot be compared across budgets because a larger $b$ is intrinsically slower even under the same runtime state.
We factor the realized latency of every profiled configuration into its nominal cost and a shared budget-normalized runtime multiplier $r_t\in\mathbb R_{>0}$:
\begin{equation}
\label{eq:load_factor}
\begin{aligned}
\ell_t(b)
&=L_0(b)r_t,
\qquad b\in\mathcal B_{\mathrm F},\\[-1pt]
r_t
&=\frac{\ell_t(b_t)}{L_0(b_t)},
\qquad
r_{t+1}\mid\mathbf c_t\sim\mathcal R_t.
\end{aligned}
\end{equation}
Here $\mathcal R_t$ denotes the short-horizon distribution of the next normalized runtime load.
The realized $r_t$ is observable from the completed forward: $r_t\simeq1$ indicates nominal latency and $r_t>1$ indicates a slower runtime regime.
We track its local level and dispersion with exponentially weighted statistics:
\begin{equation}
\label{eq:compute_state}
\begin{aligned}
e_t
&:=r_t-\mu_{t-1},\\[-1pt]
\mu_t
&=(1-\alpha)\mu_{t-1}+\alpha r_t,\\[-1pt]
\sigma_t^2
&=(1-\alpha)\sigma_{t-1}^2+\alpha e_t^2,\\[-1pt]
\mathbf c_t
&=\Phi_\alpha\!\left(\mathcal H^{\mathrm{run}}_t\right)
:=\big[\mu_t,\,\sigma_t,\,r_t\big]^\top\in\mathbb R^3,
\qquad
\mu_0=1,\ \sigma_0=0,\ \alpha\in(0,1].
\end{aligned}
\end{equation}
Thus $\mu_t$ estimates the local runtime level, while $\sigma_t$ tracks recent one-step prediction error.
Together with the latest observation $r_t$, they form a compact causal compute state without introducing a learned latency predictor.

\textbf{One-sided feasibility.}
For a target violation probability $\delta\in(0,1)$, the chance constraint in Eq.~\ref{eq:chance_objective} can be written in the normalized load domain as
$\Pr[r_{t+1}\le\tau_{\max}/L_0(b)\mid\mathbf c_t]\ge1-\delta$.
If $\mu_t$ and $\sigma_t$ equal the conditional mean and standard deviation of $r_{t+1}$, Cantelli's inequality
$\Pr[r_{t+1}-\mu_t\ge\lambda\sigma_t]\le(1+\lambda^2)^{-1}$
gives the sufficient coefficient
$\kappa_\delta=\sqrt{(1-\delta)/\delta}$.
With the online statistics in Eq.~\ref{eq:compute_state}, we retain the same one-sided form and calibrate $\kappa_\delta$ on held-out runtime traces:
\begin{equation}
\label{eq:robust_feasible}
\begin{aligned}
\widehat\ell^{\,+}_{t+1}(b\mid\mathbf c_t)
&:=L_0(b)\Big(\mu_t+\kappa_\delta\sigma_t\Big),\\[-1pt]
\mathcal F_{t+1}(\mathbf c_t)
&:=\Big\{
b\in\mathcal B_{\mathrm F}:
\widehat\ell^{\,+}_{t+1}(b\mid\mathbf c_t)
\le\tau_{\max}
\Big\}.
\end{aligned}
\end{equation}
When the conditional moments are exact and $\kappa_\delta=\sqrt{(1-\delta)/\delta}$, membership in $\mathcal F_{t+1}$ is sufficient for the original chance constraint.
With online estimates, Eq.~\ref{eq:robust_feasible} defines the operational admissible set: it contracts when either the estimated runtime load $\mu_t$ or its recent dispersion $\sigma_t$ increases.

% \textbf{Pre-inference allocation.}
% SlackDrive then executes the complete state -> allocation -> generation chain
% \begin{equation}
% \label{eq:full_policy}
% \begin{aligned}
% \mathbf c_t
% &=\Phi_\alpha\!\left(\mathcal H^{\mathrm{run}}_t\right),
% \qquad
% \mathcal F_{t+1}(\mathbf c_t)
% =\Big\{
% b\in\mathcal B_{\mathrm F}:
% L_0(b)(\mu_t+\kappa_\delta\sigma_t)\le\tau_{\max}
% \Big\},\\[-1pt]
% b_{t+1}^{\star}
% &=\pi_{\tau_{\max}}(\mathbf c_t;\mathcal P)
% :=\begin{cases}
% \displaystyle
% \arg\max_{b\in\mathcal F_{t+1}(\mathbf c_t)}U(b),
% &\mathcal F_{t+1}(\mathbf c_t)\neq\varnothing,\\[6pt]
% \displaystyle
% \arg\min_{b\in\mathcal B_{\mathrm F}}L_0(b),
% &\mathcal F_{t+1}(\mathbf c_t)=\varnothing,
% \end{cases}\\[-1pt]
% \big(\hat s_{t+2},\hat{\mathbf A}_{t+1}\big)
% &\sim
% f^{\mathrm{WA}}_\theta
% \!\left(
% \,\cdot\mid
% s_{t+1},o_{t+1},g_{t+1};
% b_{t+1}^{\star}
% \right)\\[-2pt]
% &\equiv
% \widetilde f_{\theta,\pi}
% \!\left(
% \,\cdot\mid
% s_{t+1},o_{t+1},g_{t+1},
% \mathbf c_t;
% \tau_{\max}
% \right).
% \end{aligned}
% \end{equation}
% The expensive world-action forward therefore occurs only after $b_{t+1}^{\star}$ has been chosen.
% When runtime headroom increases, Eq.~\ref{eq:full_policy} re-admits higher-utility Pareto points; under contention, it contracts the admissible set before the next forward rather than reacting after the latency envelope has already been violated.
\textbf{Pre-inference allocation.}
SlackDrive executes the complete
$\text{state}\rightarrow\text{feasibility}\rightarrow\text{allocation}\rightarrow\text{generation}$
chain
\begin{equation}
\label{eq:full_policy}
\begin{aligned}
\mathbf c_t
&=
\Phi_\alpha\!\left(\mathcal H^{\mathrm{run}}_t\right),
\\[2pt]
\mathcal F_{t+1}(\mathbf c_t)
&=
\Big\{
b\in\mathcal B_{\mathrm F}
:
L_0(b)\big(\mu_t+\kappa_\delta\sigma_t\big)
\le\tau_{\max}
\Big\},
\\[2pt]
b_{t+1}^{\star}
&=
\pi_{\tau_{\max}}(\mathbf c_t;\mathcal P)
:=
\begin{cases}
\displaystyle
\arg\max_{b\in\mathcal F_{t+1}(\mathbf c_t)} U(b),
&
\mathcal F_{t+1}(\mathbf c_t)\neq\varnothing,
\\[5pt]
\displaystyle
\arg\min_{b\in\mathcal B_{\mathrm F}} L_0(b),
&
\mathcal F_{t+1}(\mathbf c_t)=\varnothing,
\end{cases}
\\[3pt]
\big(\hat s_{t+2},\hat{\mathbf A}_{t+1}\big)
&\sim
f^{\mathrm{WA}}_\theta
\!\left(
\cdot
\mid
s_{t+1},o_{t+1},g_{t+1};
b_{t+1}^{\star}
\right)
\\[-1pt]
&\equiv
\widetilde f_{\theta,\pi}
\!\left(
\cdot
\mid
s_{t+1},o_{t+1},g_{t+1},
\mathbf c_t;
\tau_{\max}
\right).
\end{aligned}
\end{equation}
The expensive world-action forward therefore occurs only after $b_{t+1}^{\star}$ has been chosen.
When runtime headroom increases, Eq.~\ref{eq:full_policy} re-admits higher-utility Pareto points; under contention, it contracts the admissible set before the next forward rather than reacting only after the latency envelope has already been violated during execution.

\textbf{Actuator independence.}
The allocation rule requires only a discrete inference configuration whose quality and latency can be profiled.
For a token-pruned VLA, a query-based world-action model, or an iterative diffusion planner, the same abstract budget can be instantiated as
\begin{equation}
\label{eq:actuator_map}
b_t\ \longmapsto\ \xi(b_t)
=\begin{cases}
N_{\mathrm{vis}}(b_t), & \text{retained visual-token count},\\
N_{\mathrm{qry}}(b_t), & \text{active action/world queries},\\
N_{\mathrm{step}}(b_t), & \text{iterative generation steps},
\end{cases}
\qquad
\theta\ \text{fixed for all }b_t\in\mathcal B_{\mathrm F}.
\end{equation}
SlackDrive therefore changes the compute policy around the model rather than introducing a second learned model or modifying the backbone parameters.

\begin{algorithm}[H]
\caption{SlackDrive pre-inference compute allocation}
\label{alg:efficientdrive}
\begin{algorithmic}[1]
\Require profile $\mathcal P$; admissible latency $\tau_{\max}$; state rate $\alpha$; margin $\kappa_\delta$
\State $\mu\gets1$, $\sigma^2\gets0$
\For{control step $t=0,1,2,\ldots$}
    \State
    $\mathcal F_{t+1}
    \gets
    \{b\in\mathcal B_{\mathrm F}:
    L_0(b)(\mu+\kappa_\delta\sigma)\le\tau_{\max}\}$
    \State
    $b_{t+1}\gets
    \arg\max_{b\in\mathcal F_{t+1}}U(b)$
    if $\mathcal F_{t+1}\neq\emptyset$;
    otherwise choose
    $\arg\min_{b\in\mathcal B_{\mathrm F}}L_0(b)$
    \State
    Execute
    $(\hat s_{t+2},\hat{\mathbf A}_{t+1})
    \sim
    f_\theta^{\mathrm{WA}}
    (\cdot\mid s_{t+1},o_{t+1},g_{t+1};b_{t+1})$
    and measure $\ell_{t+1}$
    \State
    $r_{t+1}\gets\ell_{t+1}/L_0(b_{t+1})$;
    $\mu_{\rm old}\gets\mu$
    \State
    $\mu\gets(1-\alpha)\mu+\alpha r_{t+1}$
    \State
    $\sigma^2\gets
    (1-\alpha)\sigma^2+
    \alpha(r_{t+1}-\mu_{\rm old})^2$
\EndFor
\end{algorithmic}
\end{algorithm}

\subsection{Computational Complexity}
\label{subsec:complexity}

Let $C_f(b)$ denote the neural cost of one forward pass under budget $b$, and let
$\eta_{t,k}=\Pr[b_t=b_k]$
be the occupancy of frontier point $b_k\in\mathcal B_{\mathrm F}$ induced by the runtime process.
The controller evaluates $|\mathcal B_{\mathrm F}|$ scalar inequalities and updates a three-dimensional compute state, giving
$O(|\mathcal B_{\mathrm F}|)$ controller time per step,
$O(1)$ online state, and
$O(|\mathcal B_{\mathrm F}|)$ profile storage.
For an episode of $T$ steps, a token-budgeted Transformer with $H_L$ layers, hidden width $D$, feed-forward width $M$, and budget-dependent sequence length $S(b_k)$ has expected total cost
\begin{equation}
\label{eq:episode_cost}
\begin{aligned}
\mathbb E[C_{1:T}]
&=
\sum_{t=1}^{T}
\sum_{k=1}^{|\mathcal B_{\mathrm F}|}
\eta_{t,k}\,C_f(b_k)
+
O\!\left(T|\mathcal B_{\mathrm F}|\right),
\qquad
\sum_k\eta_{t,k}=1,\\[-1pt]
C_f(b_k)
&\approx
H_L\!\left[
2S(b_k)^2D
+
4S(b_k)D^2
+
\gamma_{\mathrm{ffn}}S(b_k)DM
\right]
+
C_{\mathrm{act}}(b_k),
\end{aligned}
\end{equation}
where $\gamma_{\mathrm{ffn}}=2$ for a standard two-layer FFN and
$\gamma_{\mathrm{ffn}}=3$ for a gated FFN under multiply accumulate counting, and
$C_{\mathrm{act}}(b_k)$ accounts for a budget-dependent action or iterative generation head when present.
The controller cost is therefore independent of the quadratic Transformer term and remains negligible for the small frontier used in practice; for non-token actuators, Eq.~\ref{eq:episode_cost} is unchanged and only $C_f(b)$ takes a different profiled form.

\section{Experiments}
\label{sec:experiments}

We evaluate whether online pre-inference allocation preserves more latency-valid planning quality than a compute configuration fixed before deployment.
Our experiments are organized around five questions: (1) does EfficientDrive improve planning utility under a strict latency constraint, (2) does it adapt across different admissible latency budgets, (3) does the allocation rule transfer to a different backbone and compute actuator, (4) which feedback terms are necessary, and (5) how sensitive is the controller to its allocation hyperparameters?

\subsection{Experimental Setup}
\label{subsec:exp_setup}

\begin{table*}[t]
\centering
\footnotesize
% \caption{\textbf{Main results on NAVSIM v2 under latency constraints.} All methods are evaluated on the same $N=4{,}966$ held-out scenes and fluctuating-load trace. \textbf{Lat.}, \textbf{FLOPs}, and \textbf{Mem.} report the realized compute profile. The nine NAVSIM terms are \textbf{NC}: No at-fault Collision, \textbf{DAC}: Drivable Area Compliance, \textbf{DDC}: Driving Direction Compliance, \textbf{TLC}: Traffic Light Compliance, \textbf{EP}: Ego Progress, \textbf{TTC}: Time to Collision, \textbf{LK}: Lane Keeping, \textbf{HC}: History Comfort, and \textbf{EC}: Extended Comfort. \textbf{EPDMS@$\tau_{\max}$} applies the latency-valid scoring rule in Eq.~\ref{eq:epdms_latency}. Underlined EPDMS@$\tau_{\max}$ values denote the strongest baseline within each latency regime; full-token rows are shaded gray, while EfficientDrive rows are highlighted in lavender and their latency-constrained scores are bold.}
\caption{\textbf{Main results on NAVSIM v2 under latency constraints.}
All methods use the same $N=4{,}966$ held-out scenes and fluctuating-load trace.
\textbf{EPDMS@$\tau_{\max}$} follows Eq.~\ref{eq:epdms_latency}.
Underlined values denote the best baseline; full-token rows are gray and EfficientDrive rows are lavender.}
\label{tab:main}
\renewcommand{\arraystretch}{1.03}
\setlength{\tabcolsep}{2.7pt}

\resizebox{\textwidth}{!}{
\begin{tabular}{l|ccc|ccccccccc|c}
\toprule

\multirow{2}{*}{\textbf{Method}}
& \multicolumn{3}{c|}{\textbf{Compute Profile}}
& \multicolumn{9}{c|}{\textbf{Latency-Valid NAVSIM Submetrics} $\uparrow$}
& \multicolumn{1}{c}{\textbf{Quality} $\uparrow$}\\

\cmidrule(lr){2-4}
\cmidrule(lr){5-13}
\cmidrule(lr){14-14}

& \textbf{Lat.}$\downarrow$
& \textbf{FLOPs}$\downarrow$
& \textbf{Mem.}$\downarrow$
& \textbf{NC}
& \textbf{DAC}
& \textbf{DDC}
& \textbf{TLC}
& \textbf{EP}
& \textbf{TTC}
& \textbf{LK}
& \textbf{HC}
& \textbf{EC}
& \textbf{EPDMS@$\tau_{\max}$}\\

& \textbf{ms/fr}
& \textbf{TF}
& \textbf{GB}
& & & & & & & & & & \\

\midrule

% \multicolumn{14}{c}{\textbf{Admissible latency $\tau_{\max}=130$ ms}}\\
% \cmidrule(lr){1-14}
% \rowcolor{softgray}Full-token & 226 & 56.9 & 11.0 & 0.00 & 0.00 & 0.00 & 0.00 & 0.00 & 0.00 & 0.00 & 0.00 & 0.00 & 0.00\\
% FastV & 188 & 47.3 & 10.4 & 0.00 & 0.00 & 0.00 & 0.00 & 0.00 & 0.00 & 0.00 & 0.00 & 0.00 & 0.00\\
% SparseVLM & 188 & 47.3 & 10.4 & 0.00 & 0.00 & 0.00 & 0.00 & 0.00 & 0.00 & 0.00 & 0.00 & 0.00 & 0.00\\
% EfficientVLA & 188 & 47.3 & 10.4 & 0.00 & 0.00 & 0.00 & 0.00 & 0.00 & 0.00 & 0.00 & 0.00 & 0.00 & 0.00\\
% ToMe & 182 & 45.7 & 10.2 & 0.00 & 0.00 & 0.00 & 0.00 & 0.00 & 0.00 & 0.00 & 0.00 & 0.00 & 0.00\\
% VisionZip & 182 & 45.7 & 10.2 & 0.00 & 0.00 & 0.00 & 0.00 & 0.00 & 0.00 & 0.00 & 0.00 & 0.00 & 0.00\\
% PruMerge+ & 182 & 45.7 & 10.2 & 0.00 & 0.00 & 0.00 & 0.00 & 0.00 & 0.00 & 0.00 & 0.00 & 0.00 & 0.00\\
% Static@432 & 84 & 14.4 & 8.0 & 59.54 & 58.04 & 61.85 & 62.17 & 55.09 & 58.98 & 59.20 & 61.21 & 46.83 & 50.70\\
% Static@135 & 54 & 4.5 & 6.4 & 84.99 & 84.53 & 96.20 & 98.85 & 88.15 & 83.93 & 91.50 & 98.14 & 65.00 & \underline{63.03}\\
% \rowcolor{lavender}\textbf{EfficientDrive} & 73 & 10.9 & 7.5 & 92.78 & 90.94 & 98.50 & 99.44 & 88.52 & 91.65 & 93.64 & 98.11 & 73.06 & \textbf{76.71}\\
% \midrule

\multicolumn{14}{c}{\textbf{Admissible latency $\tau_{\max}=180$ ms}}\\
\cmidrule(lr){1-14}

\rowcolor{softgray}Full-token
& 226 & 56.9 & 11.0
& 0.00 & 0.00 & 0.00 & 0.00 & 0.00 & 0.00 & 0.00 & 0.00 & 0.00
& 0.00\\

FastV
& 188 & 47.3 & 10.4
& 0.00 & 0.00 & 0.00 & 0.00 & 0.00 & 0.00 & 0.00 & 0.00 & 0.00
& 0.00\\

SparseVLM
& 188 & 47.3 & 10.4
& 0.00 & 0.00 & 0.00 & 0.00 & 0.00 & 0.00 & 0.00 & 0.00 & 0.00
& 0.00\\

% EfficientVLA
% & 188 & 47.3 & 10.4
% & 0.00 & 0.00 & 0.00 & 0.00 & 0.00 & 0.00 & 0.00 & 0.00 & 0.00
% & 0.00\\

ToMe
& 182 & 45.7 & 10.2
& 0.00 & 0.00 & 0.00 & 0.00 & 0.00 & 0.00 & 0.00 & 0.00 & 0.00
& 0.00\\

VisionZip
& 182 & 45.7 & 10.2
& 0.00 & 0.00 & 0.00 & 0.00 & 0.00 & 0.00 & 0.00 & 0.00 & 0.00
& 0.00\\

PruMerge+
& 182 & 45.7 & 10.2
& 0.00 & 0.00 & 0.00 & 0.00 & 0.00 & 0.00 & 0.00 & 0.00 & 0.00
& 0.00\\

Static@432
& 84 & 14.4 & 8.0
& 95.22 & 93.04 & 99.22 & 99.57 & 88.35 & 94.34 & 94.70 & 98.11 & 75.18
& \underline{81.13}\\

Static@135
& 54 & 4.5 & 6.4
& 84.99 & 84.53 & 96.20 & 98.85 & 88.15 & 83.93 & 91.50 & 98.14 & 65.00
& 63.03\\

\rowcolor{lavender}\textbf{EfficientDrive}
& 109 & 21.8 & 8.5
& 95.84 & 94.32 & 99.36 & 99.65 & 88.42 & 95.15 & 95.62 & 98.11 & 76.62
& \textbf{83.29}\\

\midrule

\multicolumn{14}{c}{\textbf{Admissible latency $\tau_{\max}=250$ ms}}\\
\cmidrule(lr){1-14}

\rowcolor{softgray}Full-token
& 226 & 56.9 & 11.0
& 23.71 & 23.36 & 23.94 & 24.10 & 21.11 & 23.59 & 23.53 & 23.68 & 19.01
& 21.27\\

FastV
& 188 & 47.3 & 10.4
& 43.13 & 41.80 & 43.73 & 44.03 & 38.61 & 42.95 & 42.56 & 43.19 & 34.32
& 37.86\\

SparseVLM
& 188 & 47.3 & 10.4
& 43.07 & 41.64 & 43.74 & 44.02 & 38.63 & 42.91 & 42.54 & 43.19 & 34.12
& 37.67\\

% EfficientVLA
% & 188 & 47.3 & 10.4
% & 43.23 & 41.68 & 43.70 & 44.02 & 38.62 & 43.06 & 42.44 & 43.18 & 33.98
% & 37.78\\

ToMe
& 182 & 45.7 & 10.2
& 45.18 & 43.58 & 46.78 & 47.28 & 41.33 & 44.93 & 45.05 & 46.49 & 34.83
& 37.93\\

VisionZip
& 182 & 45.7 & 10.2
& 45.93 & 42.85 & 46.72 & 47.30 & 41.35 & 45.69 & 45.13 & 46.44 & 35.53
& 38.17\\

PruMerge+
& 182 & 45.7 & 10.2
& 45.96 & 42.90 & 46.69 & 47.31 & 41.31 & 45.75 & 45.04 & 46.46 & 35.58
& 38.27\\

Static@432
& 84 & 14.4 & 8.0
& 95.22 & 93.04 & 99.22 & 99.57 & 88.35 & 94.34 & 94.70 & 98.11 & 75.18
& \underline{81.13}\\

Static@135
& 54 & 4.5 & 6.4
& 84.99 & 84.53 & 96.20 & 98.85 & 88.15 & 83.93 & 91.50 & 98.14 & 65.00
& 63.03\\

\rowcolor{lavender}\textbf{EfficientDrive}
& 147 & 33.0 & 9.3
& 97.16 & 95.59 & 99.42 & 99.80 & 88.30 & 96.44 & 96.77 & 98.05 & 78.23
& \textbf{85.93}\\

\midrule

\multicolumn{14}{c}{\textbf{Admissible latency $\tau_{\max}=340$ ms}}\\
\cmidrule(lr){1-14}

\rowcolor{softgray}Full-token
& 226 & 56.9 & 11.0
& 58.08 & 57.34 & 58.71 & 59.04 & 51.71 & 57.76 & 57.61 & 57.85 & 46.14
& 52.20\\

FastV
& 188 & 47.3 & 10.4
& 97.78 & 95.27 & 99.42 & 99.94 & 87.53 & 97.24 & 96.93 & 98.08 & 78.17
& \underline{86.16}\\

SparseVLM
& 188 & 47.3 & 10.4
& 97.75 & 94.88 & 99.36 & 99.92 & 87.54 & 97.22 & 96.95 & 98.07 & 77.72
& 85.75\\

% EfficientVLA
% & 188 & 47.3 & 10.4
% & 97.94 & 94.67 & 99.33 & 99.91 & 87.54 & 97.41 & 96.82 & 98.06 & 77.79
% & 85.65\\

ToMe
& 182 & 45.7 & 10.2
& 95.38 & 91.80 & 98.77 & 99.77 & 87.28 & 94.79 & 95.38 & 98.12 & 73.89
& 79.85\\

VisionZip
& 182 & 45.7 & 10.2
& 96.50 & 91.08 & 98.75 & 99.83 & 87.29 & 95.88 & 95.45 & 98.06 & 75.73
& 80.80\\

PruMerge+
& 182 & 45.7 & 10.2
& 96.61 & 91.30 & 98.77 & 99.83 & 87.20 & 96.08 & 95.43 & 98.09 & 75.73
& 81.05\\

Static@432
& 84 & 14.4 & 8.0
& 95.22 & 93.04 & 99.22 & 99.57 & 88.35 & 94.34 & 94.70 & 98.11 & 75.18
& 81.13\\

Static@135
& 54 & 4.5 & 6.4
& 84.99 & 84.53 & 96.20 & 98.85 & 88.15 & 83.93 & 91.50 & 98.14 & 65.00
& 63.03\\

\rowcolor{lavender}\textbf{EfficientDrive}
& 193 & 47.1 & 10.3
& 98.10 & 96.52 & 99.48 & 99.94 & 87.80 & 97.50 & 97.38 & 98.04 & 78.70
& \textbf{87.77}\\

\bottomrule
\end{tabular}
}
\vspace{-3mm}
\end{table*}

\textbf{Benchmark and backbone.}
We evaluate on NAVSIM v2~\cite{cao2025pseudosim} and instantiate EfficientDrive on DriveDreamer-Policy~\cite{zhou2026drivedreamerpolicy}, following the backbone used by the project implementation.
The checkpoint is frozen throughout evaluation; only the inference budget is varied before each forward pass.
Planning quality is measured by Extended Predictive Driver Model Score (EPDMS), while inference efficiency is reported with visual-token count, wall-clock latency, FLOPs, and memory.

\textbf{Evaluation protocol.}
All reproduced rows in Table~\ref{tab:main} are evaluated on the same $N=4{,}966$ NAVSIM scenes, under the same fluctuating-load trace and scoring rule.
We evaluate four admissible per-step latency budgets, $\tau_{\max}\in\{130,180,250,340\}$\,ms, and use EPDMS@$\tau_{\max}$ from Eq.~\ref{eq:epdms_latency}: a planning step that does not complete inside the admissible latency envelope contributes zero to the latency-constrained score.
The latency column reports measured per-frame execution time; FLOPs and memory follow the project's common profiling and estimation protocol and are used only for within-backbone comparison.

\textbf{Baselines.}
We compare against full-token inference, six representative efficient-inference methods applied to the same backbone, including FastV~\cite{chen2024image}, SparseVLM~\cite{zhang2024sparsevlm}, EfficientVLA~\cite{yang2025efficientvla}, ToMe~\cite{bolya2023tome}, VisionZip~\cite{yang2025visionzip}, and PruMerge+~\cite{shang2025llavaprumerge}, together with two static input-budget operating points.
The visual-token baselines reduce or merge redundant tokens within the model, whereas the static-budget baselines directly reduce the input budget to $432$ or $135$ visual tokens.
This separation is important because SlackDrive allocates the input budget before inference and can therefore reduce the cost of both visual encoding and subsequent multimodal reasoning.

\subsection{Main Results}
\label{subsec:main_results}

\textbf{Latency-constrained planning quality.}
Table~\ref{tab:main} shows that SlackDrive consistently achieves the best EPDMS@$\tau_{\max}$ across all reported latency regimes.
At $180$, $250$, and $340$\,ms, SlackDrive reaches $83.29$, $85.93$, and $87.77$, improving over the strongest baseline by $2.16$, $4.80$, and $1.61$ points, respectively.
The gain is largest at the intermediate regime, where static configurations either leave substantial compute unused or incur excessive latency under the fluctuating runtime trace.

\textbf{Adaptive use of compute headroom.}
As the admissible latency increases, SlackDrive raises its realized compute from $21.8$ to $47.1$ TFLOPs and its average latency from $109$ to $193$\,ms.
This trend shows that the controller does not collapse to a uniformly conservative configuration.
Instead, it reclaims additional runtime headroom when available and converts it into higher planning quality.

\textbf{Comparison with static and token-reduction baselines.}
At $180$\,ms, the in-model token-reduction baselines fail to produce latency-valid utility, whereas the fixed $432$-token operating point reaches $81.13$.
SlackDrive improves this to $83.29$ by adapting among profiled operating points online.
At $340$\,ms, FastV becomes the strongest baseline at $86.16$, but SlackDrive still reaches $87.77$, showing that the benefit persists even when the latency constraint is substantially relaxed.

\subsection{Ablation Study}
\label{subsec:ablation}

\begin{figure*}[t]
    \centering
    \begin{subfigure}[t]{0.245\textwidth}
        \centering
        \resizebox{0.99\linewidth}{!}{
            \includegraphics{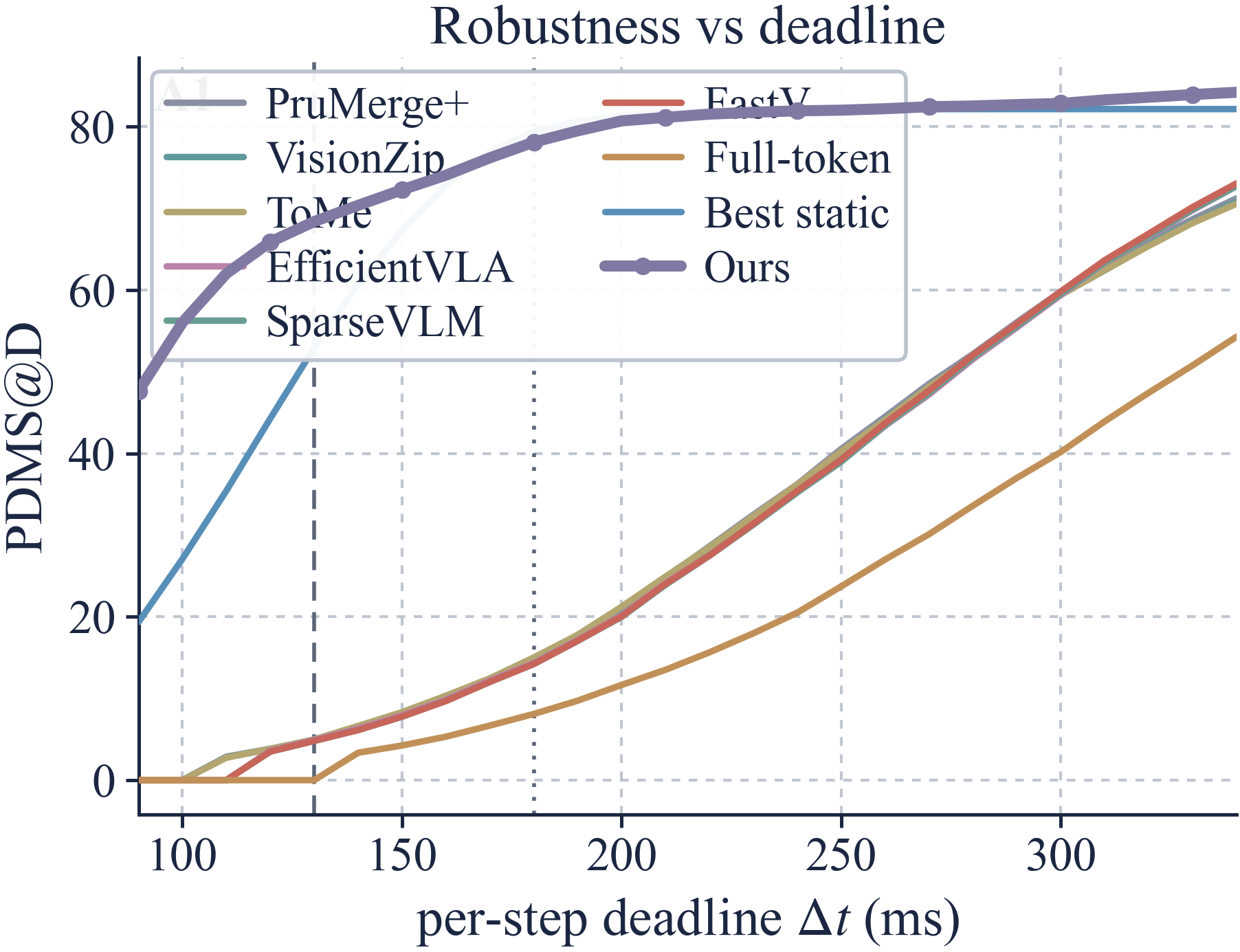}
        }
    \end{subfigure}
    \hfill
    \begin{subfigure}[t]{0.245\textwidth}
        \centering
        \resizebox{0.99\linewidth}{!}{
            \includegraphics{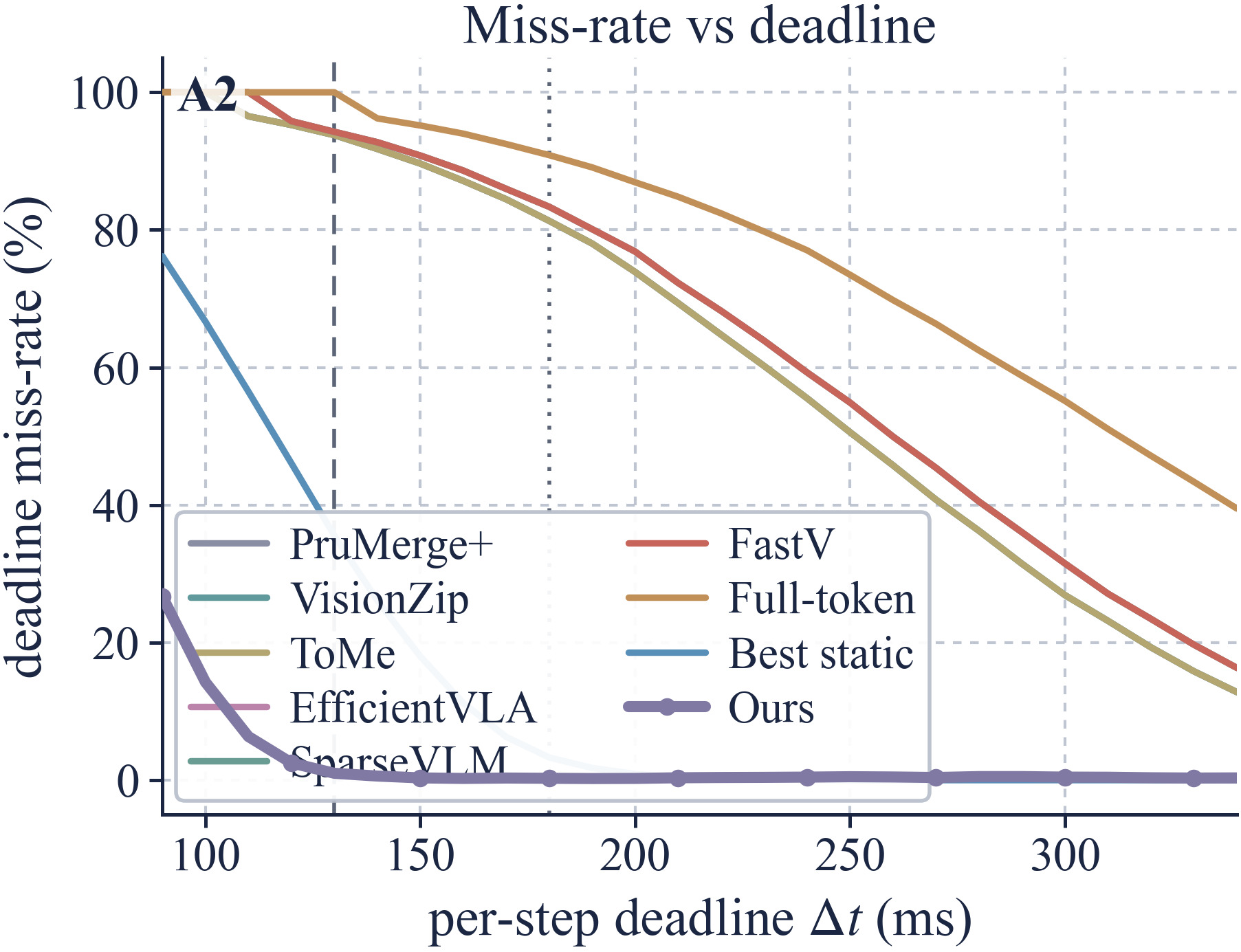}
        }
    \end{subfigure}
    \hfill
    \begin{subfigure}[t]{0.245\textwidth}
        \centering
        \resizebox{0.99\linewidth}{!}{
            \includegraphics{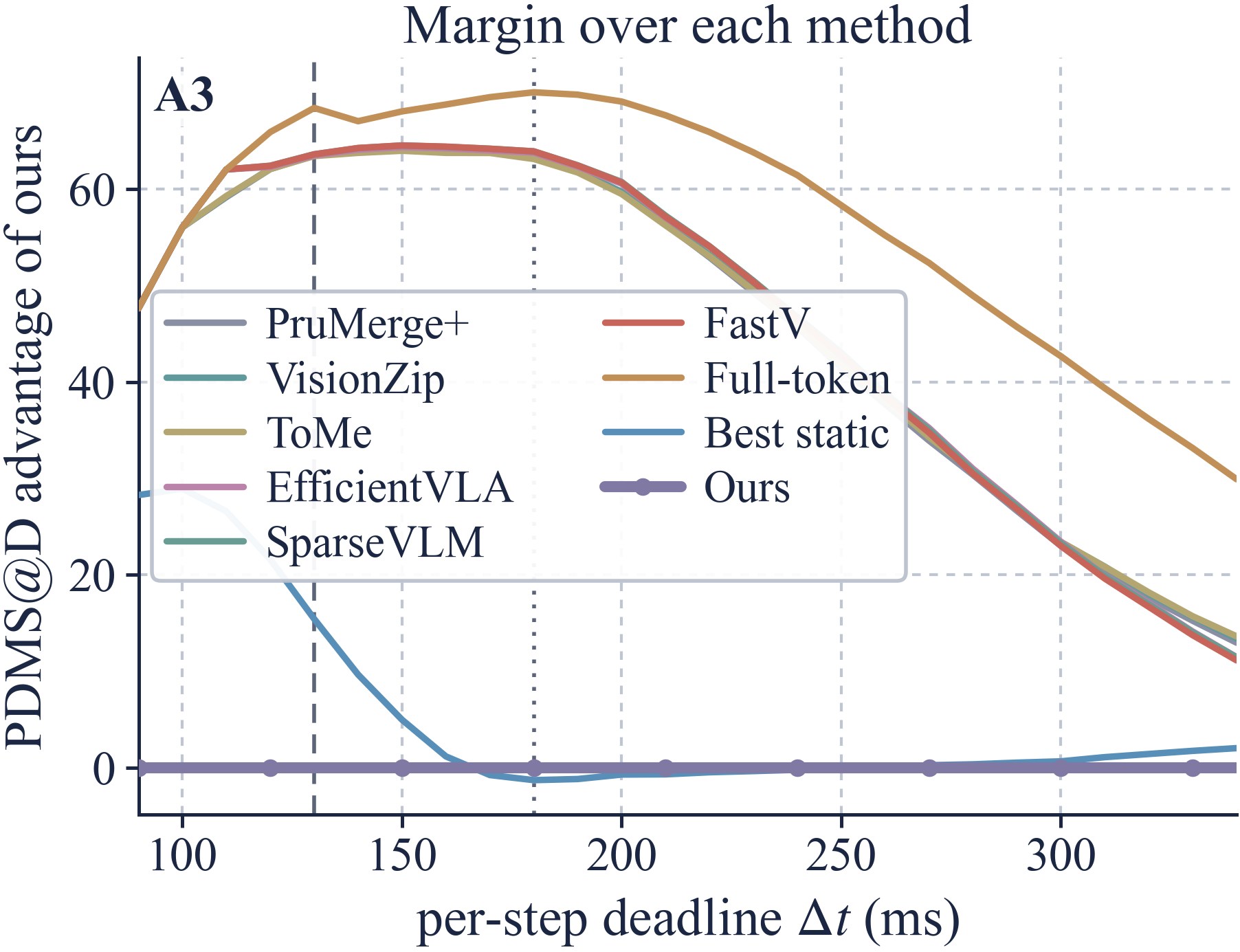}
        }
    \end{subfigure}
    \hfill
    \begin{subfigure}[t]{0.245\textwidth}
        \centering
        \resizebox{0.99\linewidth}{!}{
            \includegraphics{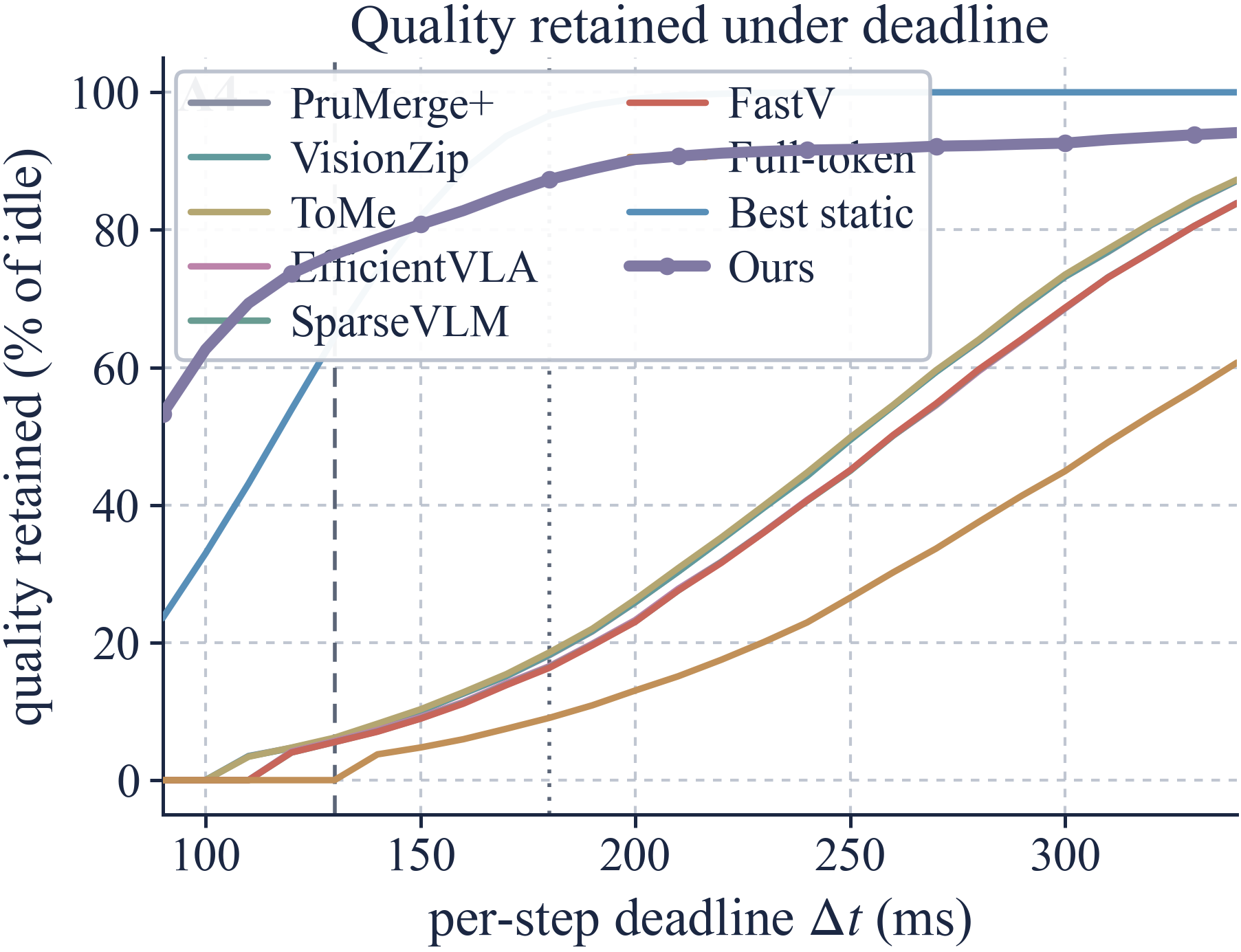}
        }
    \end{subfigure}

    \vspace{-1mm}
    \caption{\textbf{Sensitivity to the latency constraint.}
    From left to right: latency-valid utility, violation rate, gain over fixed budgets, and retained utility.}
    \vspace{-2mm}
    \label{fig:abl_latency}
\end{figure*}
\begin{figure*}[t]
    \centering
    \begin{subfigure}[t]{0.245\textwidth}
        \centering
        \resizebox{0.99\linewidth}{!}{
            \includegraphics{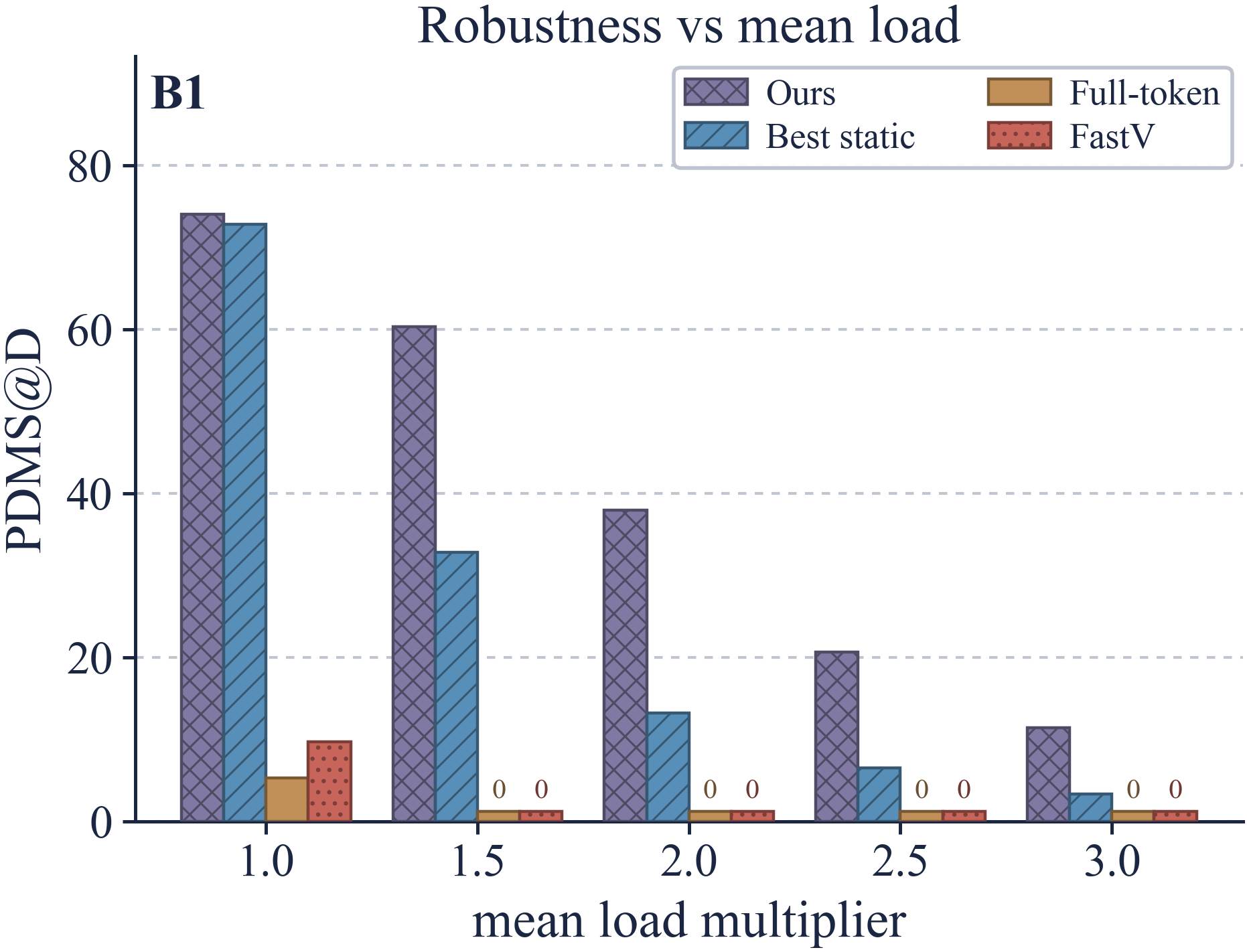}
        }
    \end{subfigure}
    \hfill
    \begin{subfigure}[t]{0.245\textwidth}
        \centering
        \resizebox{0.99\linewidth}{!}{
            \includegraphics{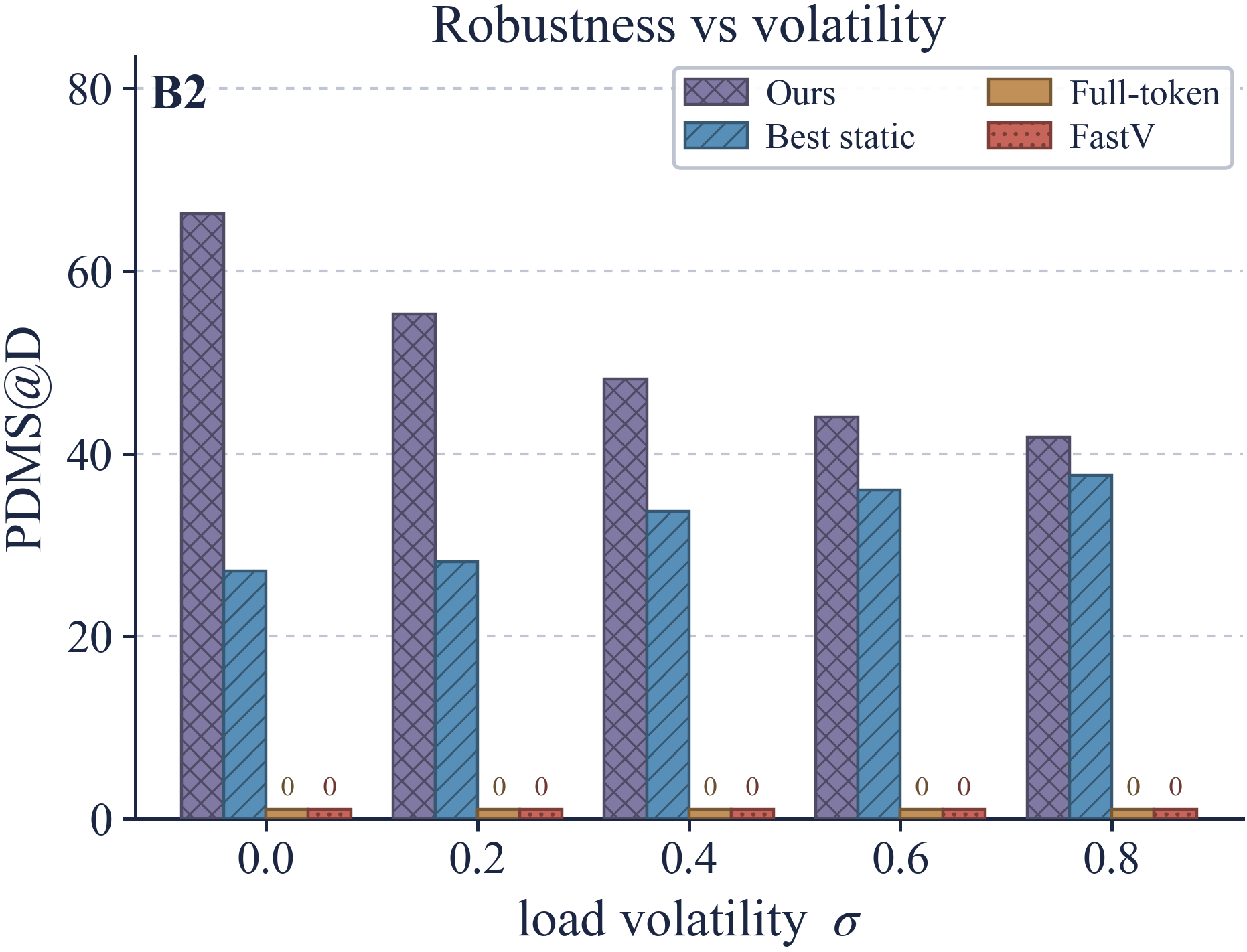}
        }
    \end{subfigure}
    \hfill
    \begin{subfigure}[t]{0.245\textwidth}
        \centering
        \resizebox{0.99\linewidth}{!}{
            \includegraphics{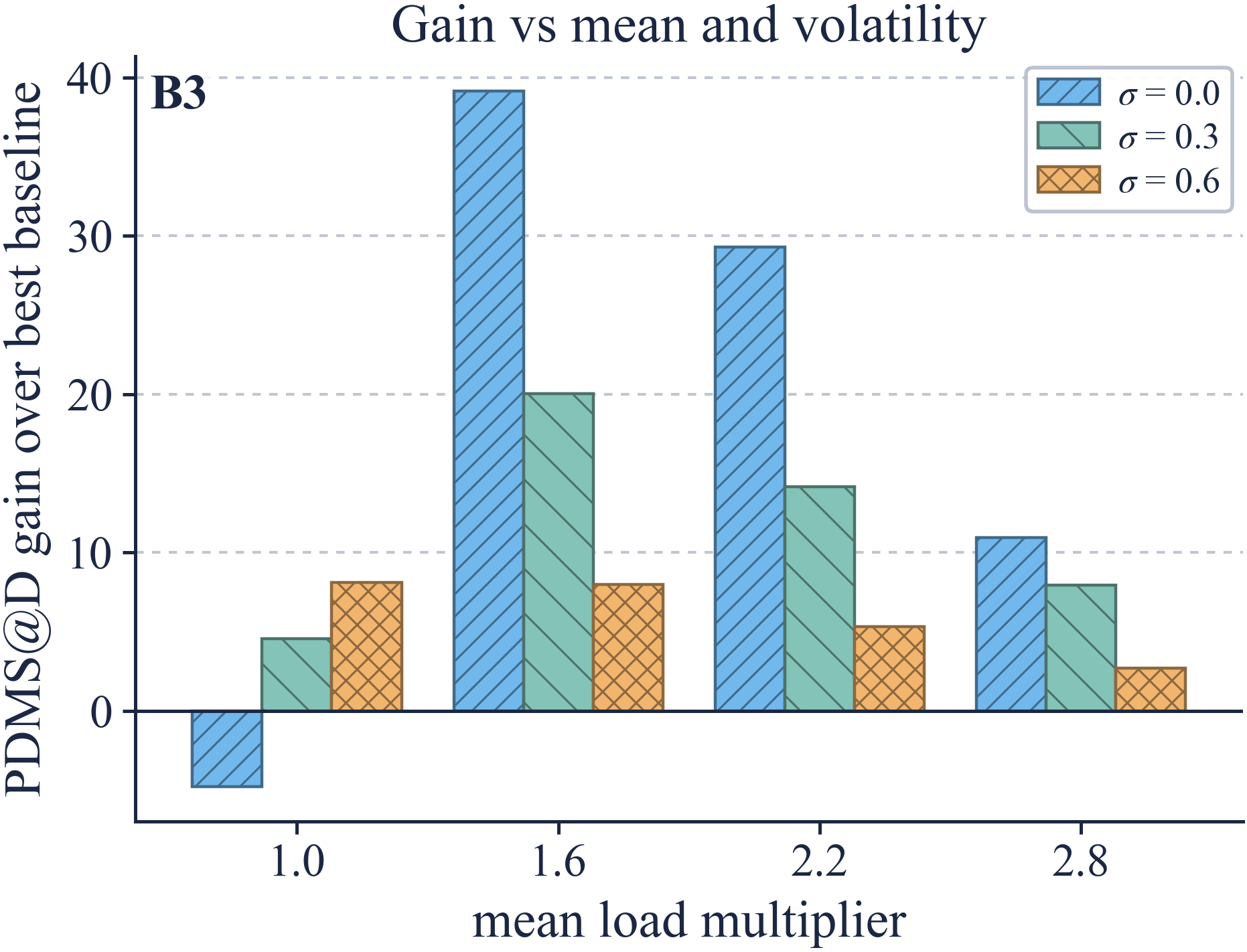}
        }
    \end{subfigure}
    \hfill
    \begin{subfigure}[t]{0.245\textwidth}
        \centering
        \resizebox{0.99\linewidth}{!}{
            \includegraphics{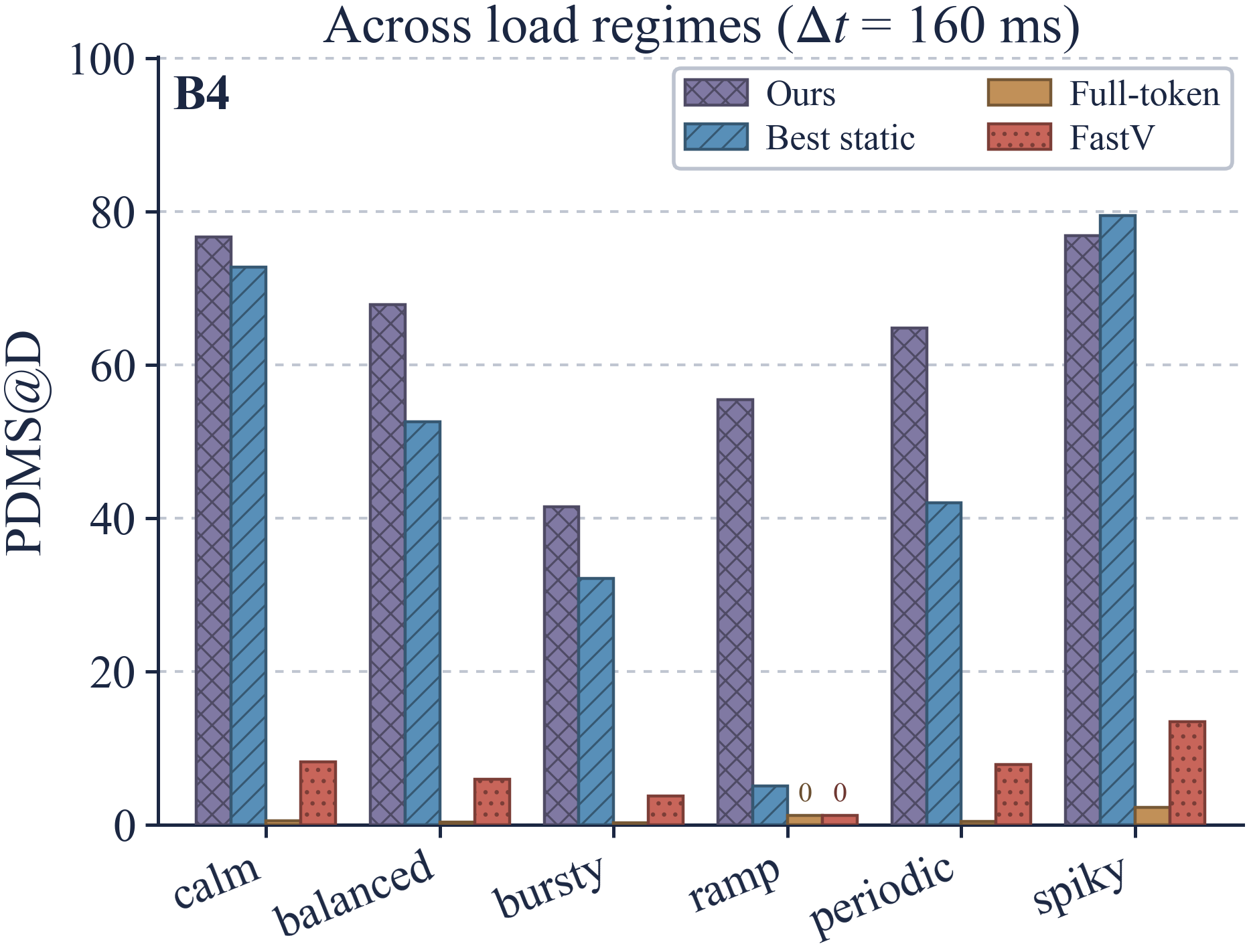}
        }
    \end{subfigure}

    \vspace{-1mm}
    \caption{\textbf{Sensitivity to runtime conditions.}
    From left to right: mean load, load variability, their joint effect, and six representative runtime regimes.}
    \vspace{-4mm}
    \label{fig:abl_load}
\end{figure*}

We isolate each factor in Eq.~\ref{eq:full_policy} while keeping the frozen world-action model and all remaining settings fixed.
These sweeps use the same latency indicator, allocation rule, and runtime model throughout.
Together, they examine when latency feedback is useful, which runtime statistics matter, and which components carry the gain.

\textbf{Latency constraint.}
We first examine how the admissible latency envelope changes the compute configuration that should be selected at inference time.
As $\tau_{\max}$ varies from 90 to 340\,ms in Fig.~\ref{fig:abl_latency}, the preferred operating point shifts substantially rather than remaining fixed across latency regimes.
At 130\,ms, adaptive allocation reaches 68.39 compared with 52.95 for the best fixed budget, while full-token inference yields zero latency-valid utility.
The violation rate remains below 1\% for $\tau_{\max}\ge130$\,ms, but rises sharply under tighter constraints because even the cheapest profiled configuration becomes infeasible.
This transition identifies an important operating boundary: below it, violations are caused by an empty feasible set rather than by an incorrect allocation decision.

\textbf{Runtime condition.}
We next investigate the runtime conditions under which recent latency provides a sufficiently informative signal for adaptive compute allocation.
To separate the effects of sustained contention from transient volatility, Fig.~\ref{fig:abl_load} independently varies the mean and variability of the normalized runtime load.
The gain over the best fixed budget increases with mean load, reaching $+8.10$ at $3\times$ load, and remains positive in five of six representative runtime regimes.
By contrast, the gain decreases as variability increases and becomes negative under memoryless spikes, where past latency provides little information about the next forward.
These results suggest that latency feedback is most valuable when runtime variation exhibits temporal structure; purely unpredictable volatility reduces forecastability without creating reliably exploitable compute headroom.

% \begin{figure*}[t]
%     \centering
%     \begin{subfigure}[t]{0.245\textwidth}
%         \centering
%         \resizebox{0.99\linewidth}{!}{
%             \includegraphics{figs/ablation2_load_p1_mean.png}
%         }
%     \end{subfigure}
%     \hfill
%     \begin{subfigure}[t]{0.245\textwidth}
%         \centering
%         \resizebox{0.99\linewidth}{!}{
%             \includegraphics{figs/ablation2_load_p2_volatility.png}
%         }
%     \end{subfigure}
%     \hfill
%     \begin{subfigure}[t]{0.245\textwidth}
%         \centering
%         \resizebox{0.99\linewidth}{!}{
%             \includegraphics{figs/ablation2_load_p3_gain_plane.png}
%         }
%     \end{subfigure}
%     \hfill
%     \begin{subfigure}[t]{0.245\textwidth}
%         \centering
%         \resizebox{0.99\linewidth}{!}{
%             \includegraphics{figs/ablation2_load_p4_regimes.png}
%         }
%     \end{subfigure}

%     \vspace{-1mm}
%     \caption{\textbf{Sensitivity to runtime conditions.}
%     From left to right: mean load, load variability, their joint effect, and six representative runtime regimes.}
%     \label{fig:abl_load}
% \end{figure*}

\begin{figure*}[t]
    \centering
    \begin{subfigure}[t]{0.245\textwidth}
        \centering
        \resizebox{0.99\linewidth}{!}{
            \includegraphics{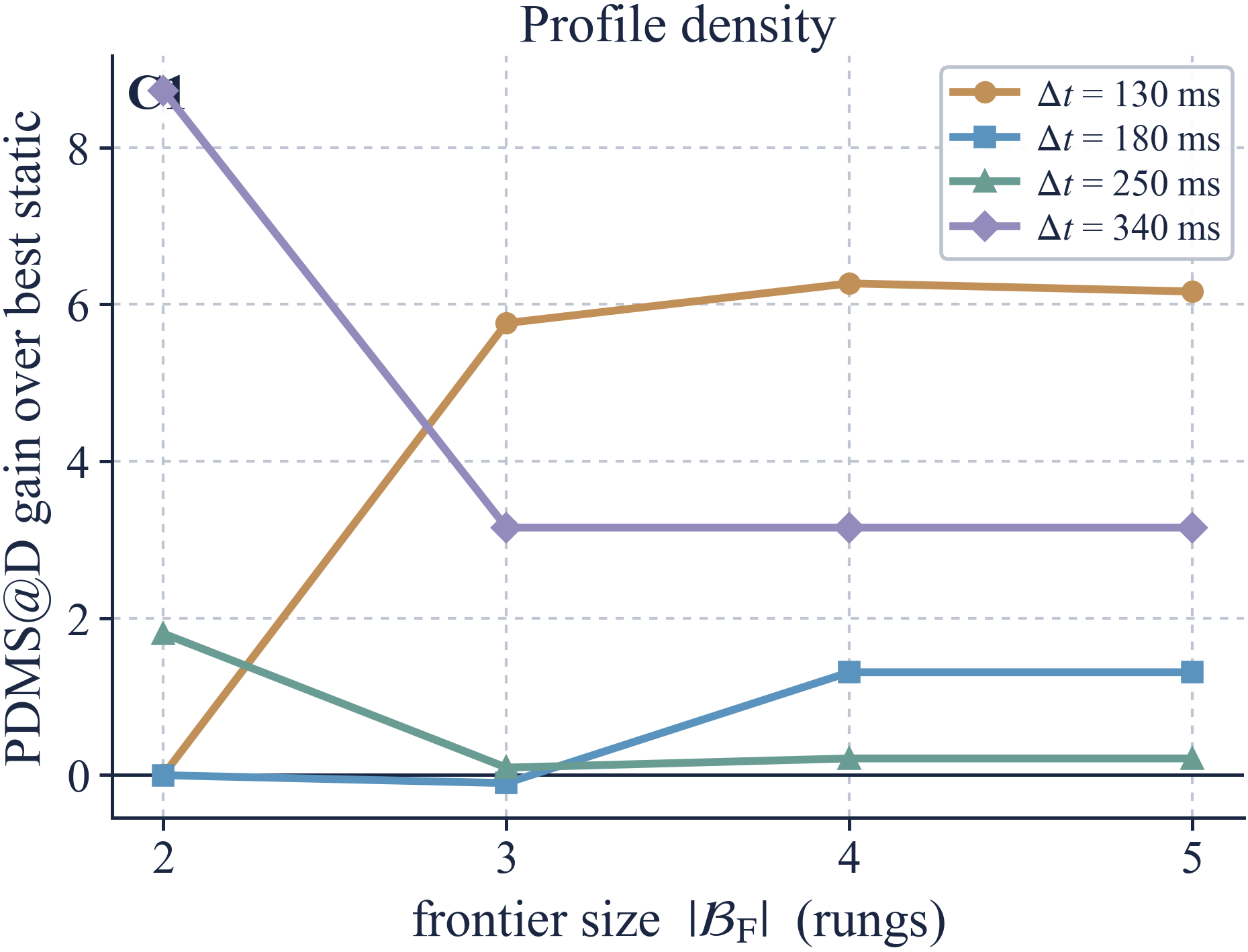}
        }
    \end{subfigure}
    \hfill
    \begin{subfigure}[t]{0.245\textwidth}
        \centering
        \resizebox{0.99\linewidth}{!}{
            \includegraphics{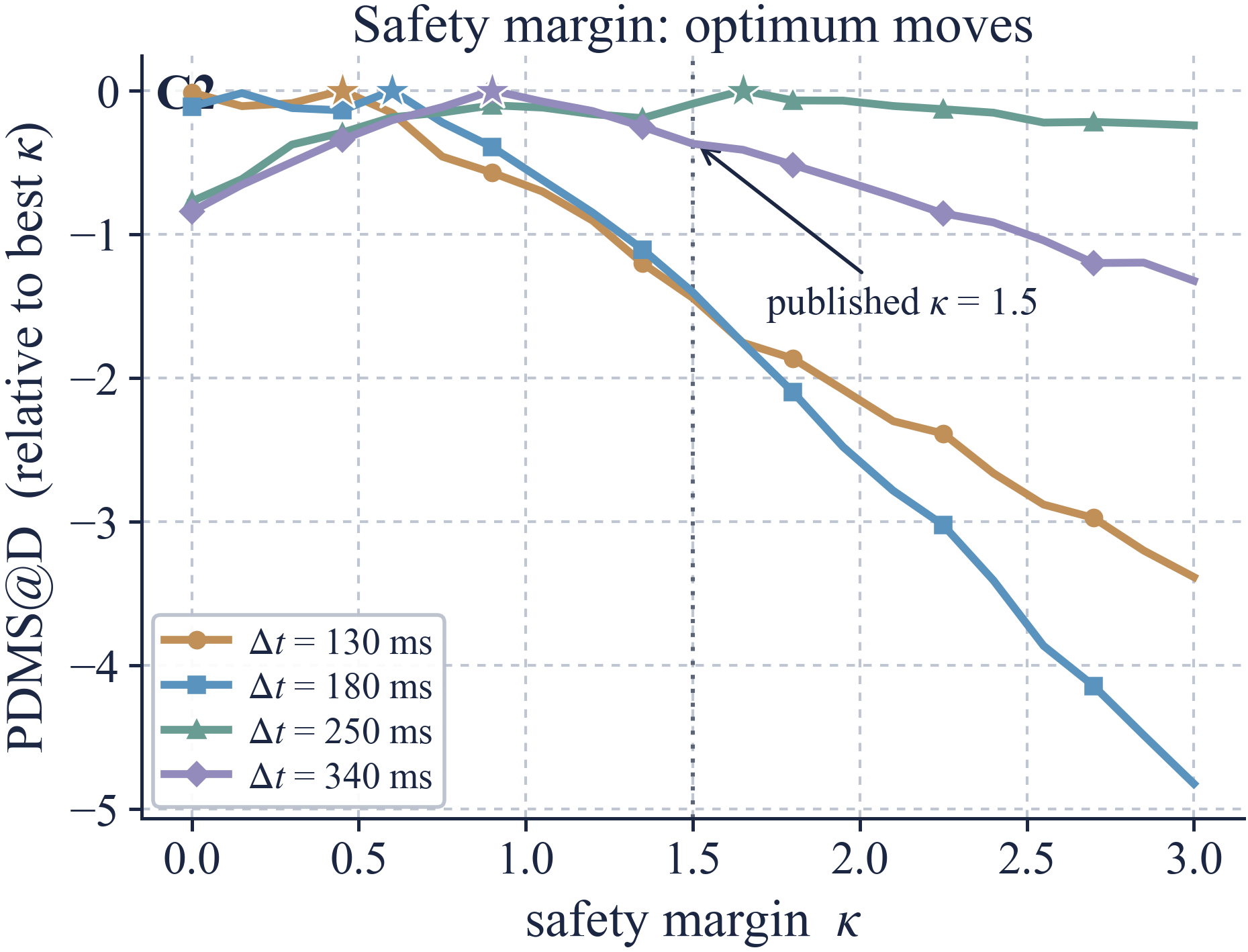}
        }
    \end{subfigure}
    \hfill
    \begin{subfigure}[t]{0.245\textwidth}
        \centering
        \resizebox{0.99\linewidth}{!}{
            \includegraphics{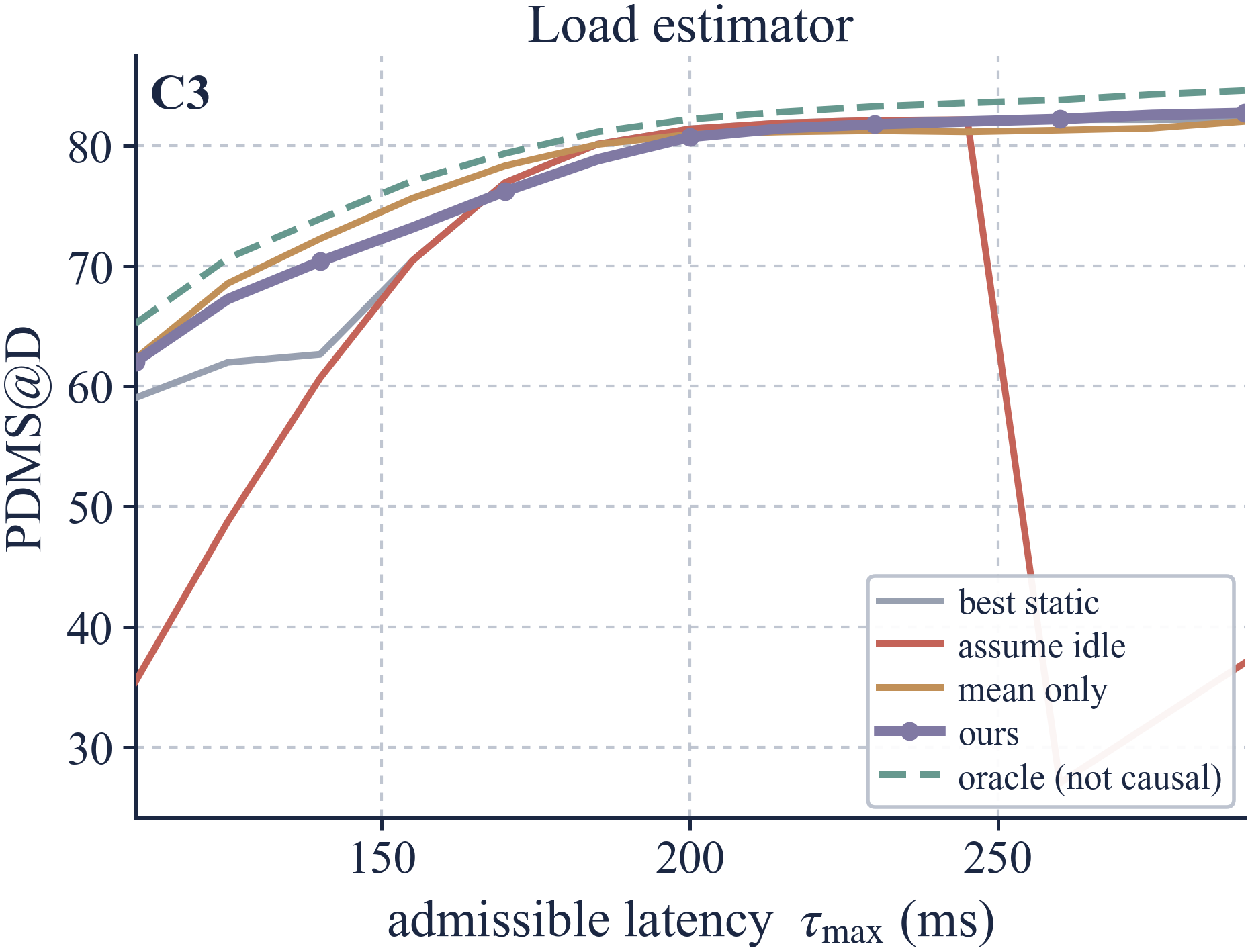}
        }
    \end{subfigure}
    \hfill
    \begin{subfigure}[t]{0.245\textwidth}
        \centering
        \resizebox{0.99\linewidth}{!}{
            \includegraphics{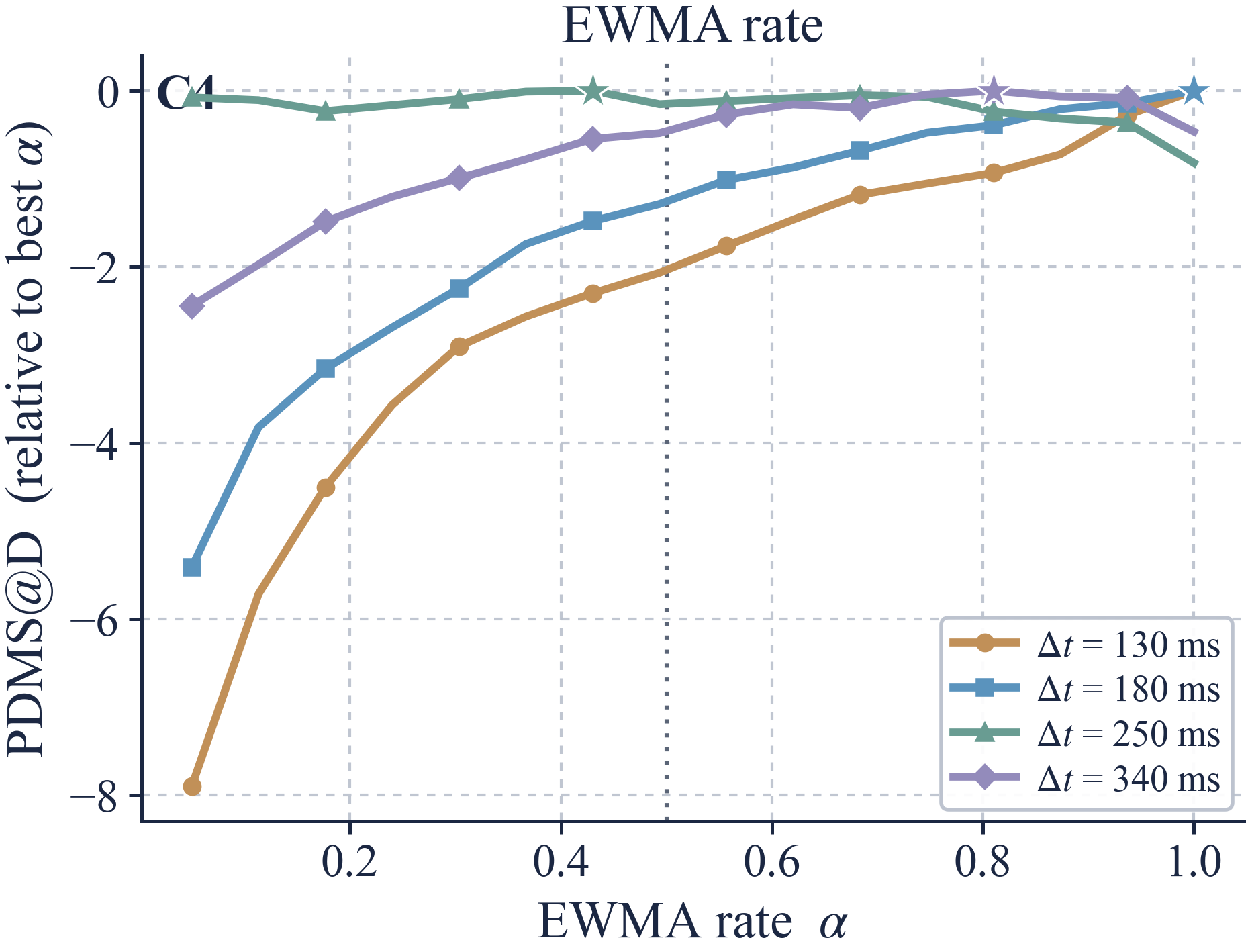}
        }
    \end{subfigure}

    \vspace{-1mm}
    \caption{\textbf{Controller design.}
    From left to right: profile density, uncertainty margin, compute-state estimator, and EWMA rate.}
    \vspace{-4mm}
    \label{fig:abl_controller}
\end{figure*}

\textbf{Controller design.}
We further ablate how SlackDrive converts observed runtime slack into additional compute.
Fig.~\ref{fig:abl_controller} evaluates profile density, uncertainty margin $\kappa_\delta$, compute-state estimation, and EWMA rate $\alpha$.
Both $\kappa_\delta$ and $\alpha$ vary with $\tau_{\max}$, showing that the controller should be calibrated to the operating envelope rather than using a single global setting.
At 140\,ms, ignoring runtime state reduces utility to 60.69, below 62.65 without scheduling, whereas the full estimator reaches 70.37 and the non-causal oracle reaches 73.91.
Mean-only estimation reaches 72.25, slightly above the full estimator, suggesting that the variance margin improves robustness but can become overly conservative under tight constraints.
The small gap to the oracle indicates that recent latency history already captures most useful runtime information.

% \begin{wraptable}{r}{0.55\columnwidth}
%     \vspace{1mm}
%     \centering
%     \scriptsize
%     \setlength{\tabcolsep}{2.0pt}
%     \renewcommand{\arraystretch}{1.12}
%     \caption{\textbf{Component removal.}
%     $\Delta$ is the utility lost relative to the full controller at each $\tau_{\max}$ (ms); negative means the removal helped.}
%     \label{tab:abl_component}

%     \begin{tabular*}{0.94\linewidth}{
%         @{\extracolsep{\fill}} l cccc rrrr @{}
%     }
%     \toprule
%     & \multicolumn{4}{c}{Retained}
%     & \multicolumn{4}{c}{$\Delta$} \\
%     \cmidrule(lr){2-5}\cmidrule(l){6-9}
%     & $\pi$ & $\Phi_\alpha$ & $\sigma_t$ & $\mathcal{B}_{\mathrm F}$
%     & 130 & 180 & 250 & 340 \\
%     \midrule

%     Full
%     & \cmark & \cmark & \cmark & \cmark
%     & \textemdash & \textemdash & \textemdash & \textemdash \\

%     $-\,\sigma_t$
%     & \cmark & \cmark & \xmark & \cmark
%     & -1.4 & -1.3 & +0.7 & +0.5 \\

%     $-\,\Phi_\alpha$
%     & \cmark & \xmark & \xmark & \cmark
%     & \textbf{+15.4} & -1.3 & -0.1 & \textbf{+29.9} \\

%     $-\,\mathcal{B}_{\mathrm F}$
%     & \cmark & \cmark & \cmark & \xmark
%     & \textbf{+6.1} & \textbf{+15.0} & \textbf{+17.6} & \textbf{+13.6} \\

%     $-\,\pi$
%     & \xmark & \xmark & \xmark & \cmark
%     & \textbf{+6.1} & -1.3 & -0.1 & \textbf{+2.1} \\

%     \bottomrule
%     \end{tabular*}
%     \vspace{-5mm}
% \end{wraptable}

\begin{figure*}[t]
    \vspace{-2mm}
    \centering

    % ===================== Main content =====================
    \makebox[\textwidth][c]{%
        % ---------- (a) Case study ----------
        \begin{minipage}[t]{0.42\textwidth}
            \vspace{0pt}
            \centering

            \begin{minipage}[c][5.3cm][c]{\linewidth}
                \centering
                \includegraphics[
                    height=4.4cm,
                    keepaspectratio
                ]{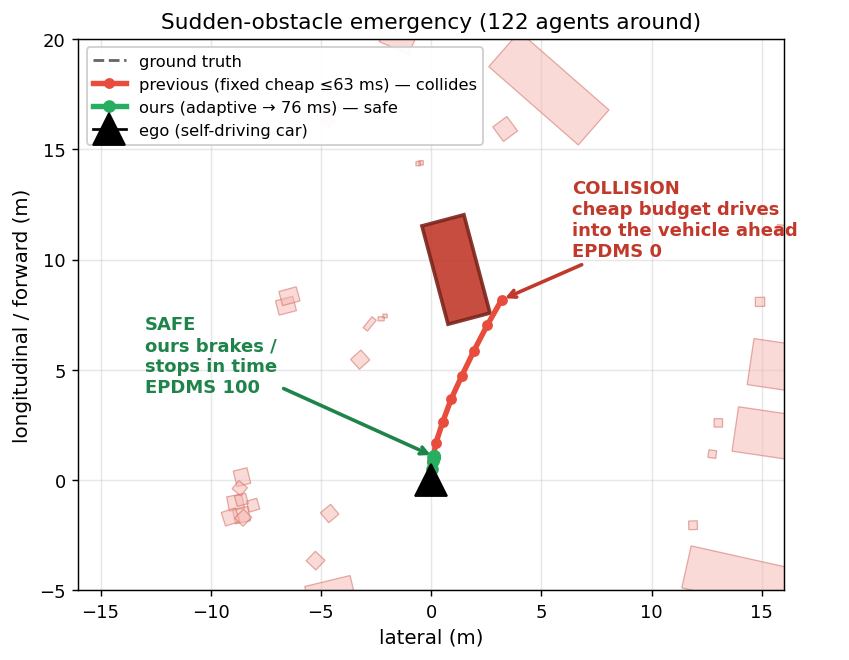}
            \end{minipage}
        \end{minipage}%
        \hspace{0.02\textwidth}%
        % ---------- (b) Component removal ----------
        \begin{minipage}[t]{0.54\textwidth}
            \vspace{0pt}
            \centering

            \begin{minipage}[c][5.3cm][c]{\linewidth}
                \centering
                \scriptsize

                \setlength{\tabcolsep}{2.3pt}
                \renewcommand{\arraystretch}{1.70}

                \begin{tabular*}{0.96\linewidth}{
                    @{\extracolsep{\fill}} l cccc rrrr @{}
                }
                \toprule
                & \multicolumn{4}{c}{Retained}
                & \multicolumn{4}{c}{$\Delta$} \\
                \cmidrule(lr){2-5}
                \cmidrule(l){6-9}

                & $\pi$
                & $\Phi_\alpha$
                & $\sigma_t$
                & $\mathcal{B}_{\mathrm F}$
                & 130
                & 180
                & 250
                & 340 \\
                \midrule

                Full
                & \cmark & \cmark & \cmark & \cmark
                & \textemdash & \textemdash & \textemdash & \textemdash \\

                $-\,\sigma_t$
                & \cmark & \cmark & \xmark & \cmark
                & -1.4 & -1.3 & +0.7 & +0.5 \\

                $-\,\Phi_\alpha$
                & \cmark & \xmark & \xmark & \cmark
                & \textbf{+15.4} & -1.3 & -0.1 & \textbf{+29.9} \\

                $-\,\mathcal{B}_{\mathrm F}$
                & \cmark & \cmark & \cmark & \xmark
                & \textbf{+6.1} & \textbf{+15.0}
                & \textbf{+17.6} & \textbf{+13.6} \\

                $-\,\pi$
                & \xmark & \xmark & \xmark & \cmark
                & \textbf{+6.1} & -1.3 & -0.1 & \textbf{+2.1} \\

                \bottomrule
                \end{tabular*}
            \end{minipage}
        \end{minipage}%
    }

    % ===================== Aligned panel captions =====================
    \vspace{-1mm}

    \makebox[\textwidth][c]{%
        \begin{minipage}[t]{0.42\textwidth}
            \centering
            {\small\bfseries
            (a) Ours avoids a critical failure}
        \end{minipage}%
        \hspace{0.02\textwidth}%
        \begin{minipage}[t]{0.54\textwidth}
            \centering
            {\small\bfseries
            (b) Component removal}
        \end{minipage}%
        }

    \vspace{-1mm}

    % \caption{
    % \textbf{Runtime adaptation and component analysis.}
    % \textbf{(a)} In a sudden-obstacle scene, a fixed low-compute configuration collides with the vehicle ahead, while SlackDrive uses available runtime headroom to select a higher compute budget and avoids the collision.
    % \textbf{(b)} Removing individual components of Eq.~\ref{eq:full_policy} shows that the resolved Pareto profile provides the most consistent gain, while recent runtime state is critical when the admissible envelope is strongly binding.
    % }
    \caption{
    \textbf{Runtime-aware allocation in practice.}
    \textbf{(a)} SlackDrive converts available runtime headroom into additional compute and avoids a collision.
    \textbf{(b)} Component removal identifies the profile and runtime state that enable this adaptation.
    }
    \vspace{-3mm}
    \label{fig:case_component}
\end{figure*}

\textbf{Component analysis and driving case.}
Finally, we examine which components drive the gain and how adaptive compute affects an individual driving decision.
In Fig.~\ref{fig:case_component}(a), a fixed low-compute configuration collides with the vehicle ahead, whereas SlackDrive uses available runtime headroom to select a higher compute budget and avoids the collision.
Fig.~\ref{fig:case_component}(b) shows that a sufficiently resolved Pareto profile provides the most consistent gain across latency regimes, while compute-state estimation becomes critical when the admissible envelope is strongly binding.
Removing the volatility term slightly improves the two tightest regimes, consistent with the conservative margin observed in Fig.~\ref{fig:abl_controller}.
Overall, the primary gain comes from selecting among well-resolved operating points using recent latency state, while the variance term controls feasible-set conservatism.

% \section{Conclusion}
% \label{sec:conclusion}

% We presented \textbf{SlackDrive}, a pre-inference compute allocator that combines a one-time quality and latency profile with online compute-state estimation for driving world-action models.
% SlackDrive improves latency-constrained planning by adapting compute to residual runtime variation while keeping the underlying model frozen.
% Experiments on NAVSIM v2 and cross-backbone transfer show that the same principle extends across different compute actuators, although the gain becomes limited when additional computation offers little utility headroom.
% Our results show that inference compute should be treated as a runtime decision variable rather than a deployment-time constant.

\section{Conclusion}
\label{sec:conclusion}

We presented \textbf{SlackDrive}, a pre-inference compute allocator that combines a one-time quality and latency profile with online compute-state estimation for driving world-action models.
SlackDrive improves latency-constrained planning by adapting compute to residual runtime variation while keeping the underlying model frozen.
Experiments on NAVSIM v2 and cross-backbone transfer show that the same principle extends across different compute actuators, although the gain becomes limited when additional computation offers little utility headroom.
This enables a single deployed model to trade computation for latency online as runtime conditions evolve.
Our results show that inference compute should be treated as a runtime decision variable rather than a deployment-time constant.

% \section{Conclusion}
% \label{sec:conclusion}

% We presented \textbf{SlackDrive}, a pre-inference compute allocator that combines a one-time quality and latency profile with online compute-state estimation for driving world-action models.
% SlackDrive improves latency-constrained planning by adapting compute to residual runtime variation while keeping the underlying model frozen.
% Experiments on NAVSIM v2 and cross-backbone transfer show that the same principle extends across different compute actuators, although the gain becomes limited when additional computation offers little utility headroom.
% By separating offline profiling from lightweight online allocation, SlackDrive provides a practical way to exploit otherwise unused runtime slack without modifying model parameters or retraining the backbone.
% This design also makes the allocator compatible with heterogeneous deployment conditions, where available latency can vary substantially across time and hardware.
% Our results show that inference compute should be treated as a runtime decision variable rather than a deployment-time constant.

% ============================================================
% References
% ============================================================

\bibliography{iclr2027_conference}
\bibliographystyle{iclr2027_conference}

% ============================================================
% Appendix
% ============================================================

\clearpage
\appendix

\section{Additional Details}
\label{sec:appendix_details}

\subsection{Reproducibility Checklist}
\label{sec:reproducibility}

The final experimental release should report the following items together with the raw runtime logs: GPU model and power mode; CUDA, PyTorch, and inference-engine versions; batch size; numerical precision; warm-up iterations; clock-control policy; competing workload used to induce runtime variation; number of latency repetitions; the exact budget grid $\mathcal B$; profile latencies $L_0(b)$; profile utilities $U(b)$; controller parameter $\alpha$; risk level $\delta$; and all random seeds.

\subsection{Static Budget Case Analysis}
\label{sec:appendix_cases}

\begin{figure*}[t]
    \centering
    \includegraphics[width=\textwidth]{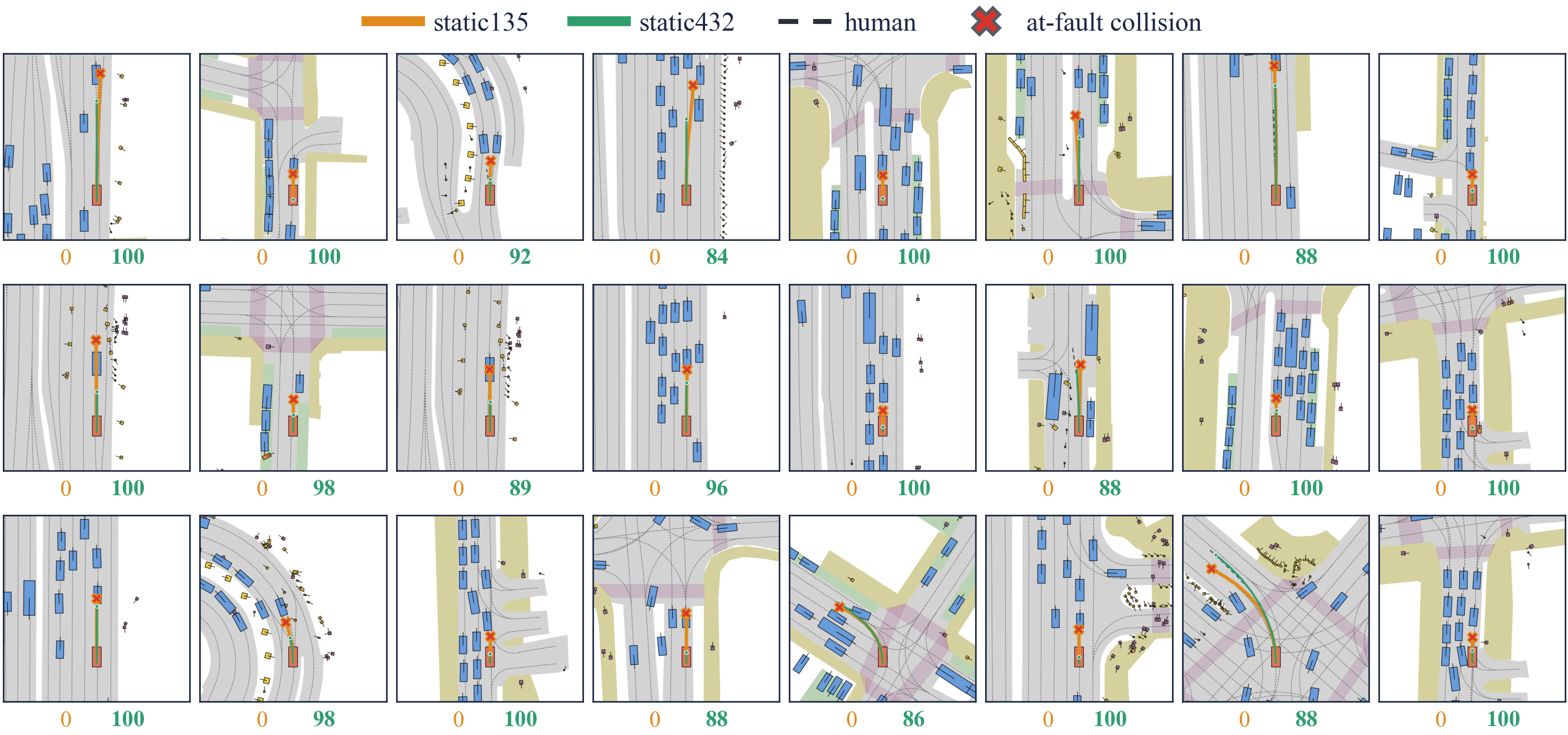}
    \caption{
    \textbf{The cheapest configuration is not always sufficient.}
    Shown are $24$ fixed-seed samples from $1{,}052$ scenes in which Static@135 collides while Static@432 remains collision-free with EPDMS $\ge80$.
    The comparison motivates selecting the highest-utility configuration inside the feasible set rather than always minimizing compute.
    }
    \label{fig:case_cheapest}
    \vspace{-5mm}
\end{figure*}

\begin{figure*}[t]
    \centering
    \includegraphics[width=\textwidth]{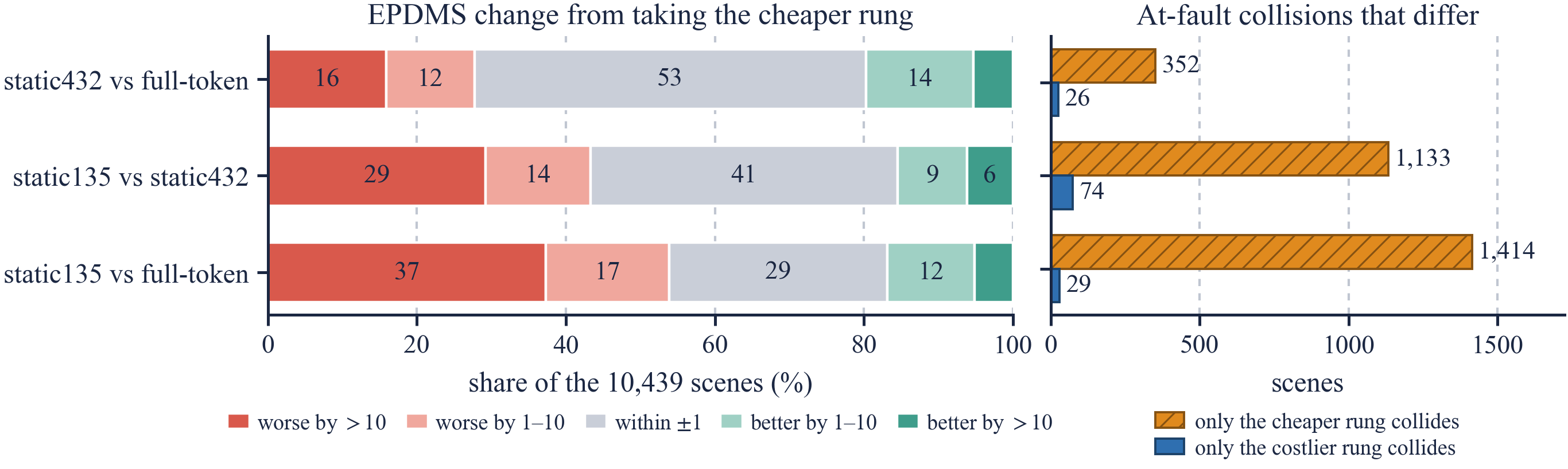}
    \caption{
    \textbf{Population structure of static compute budgets.}
    All $10{,}439$ navtest scenes are included.
    The left panel shows the per-scene EPDMS difference between Static@432 and full-token inference.
    The right panel reports collision outcomes for each pair of operating points.
    Population-level utility increases with compute, while individual scenes need not follow the same ordering.
    }
    \label{fig:case_population}
\end{figure*}

All case analyses in this section are computed from $10{,}439$ NAVSIM-v2 navtest scenes using three static operating points: full-token inference with $L_0=226$\,ms, Static@432 with $L_0=84$\,ms, and Static@135 with $L_0=54$\,ms.
Each operating point is evaluated independently without the online SlackDrive controller.
Unless otherwise specified, each gallery contains either the complete set of scenes satisfying its selection rule or a fixed-seed uniform sample from that set.
% The complete scene indices and scores are provided in \texttt{scenes.json}, together with the selection rules in \texttt{select\_scenes.py}.

\textbf{The aggregate utility frontier is monotonic even though individual scenes are not.}
Fig.~\ref{fig:case_population} shows the population structure underlying the profile utility $U(b)$ in Eq.~\ref{eq:pareto_profile}.
The three operating points achieve mean EPDMS values of $62.88$, $81.18$, and $88.48$, respectively, so utility increases monotonically with nominal latency at the population level.
However, the per-scene ordering is not monotonic: when moving from Static@432 to full-token inference, $53\%$ of scenes change by less than one EPDMS point, $28\%$ favor full-token inference, and $19\%$ favor Static@432.
The collision statistics show the same asymmetry, with substantially more scenes benefiting from additional compute than being degraded by it.
This distinction motivates profiling utility rather than equating raw computation with planning quality.

This population view also clarifies the direction of SlackDrive.
Conventional efficient inference chooses a compute configuration first and observes its latency afterward.
SlackDrive instead conditions the next configuration on the admissible latency and recent runtime state.
For example, when an unexpected vehicle cuts into the ego lane, the relevant question is not whether a predetermined token count eventually runs fast enough, but which profiled configuration can still complete within the current latency envelope while preserving the highest available planning utility.

\textbf{The cheapest feasible configuration can discard planning utility that remains affordable.}
Fig.~\ref{fig:case_cheapest} collects scenes in which Static@135 incurs an at-fault collision while Static@432 remains collision-free with EPDMS at least $80$.
There are $1{,}052$ such scenes, from which we uniformly sample $24$ using a fixed random seed.
These cases explain the first branch of Eq.~\ref{eq:full_policy}: when several configurations remain in $\mathcal F_{t+1}$, the policy selects the one with the highest profiled utility rather than the configuration with the smallest nominal latency.
Across these scenes, reducing the visual-token budget from $432$ to $135$ preserves substantial latency headroom but can remove information needed for a safe trajectory.

The same cases also illustrate the cost of the fallback branch in Eq.~\ref{eq:full_policy}.
When $\mathcal F_{t+1}$ becomes empty, SlackDrive selects the fastest profiled configuration because no candidate is predicted to satisfy the latency envelope.
This fallback protects latency as far as the available profile permits, but it cannot recover planning information that the smallest budget no longer preserves.
A sudden braking event provides the corresponding intuition: once the admissible latency becomes too restrictive, completing a smaller forward is preferable to missing the control window entirely, although the resulting plan may carry less information.

\begin{figure*}[t]
    \centering
    \includegraphics[width=\textwidth]{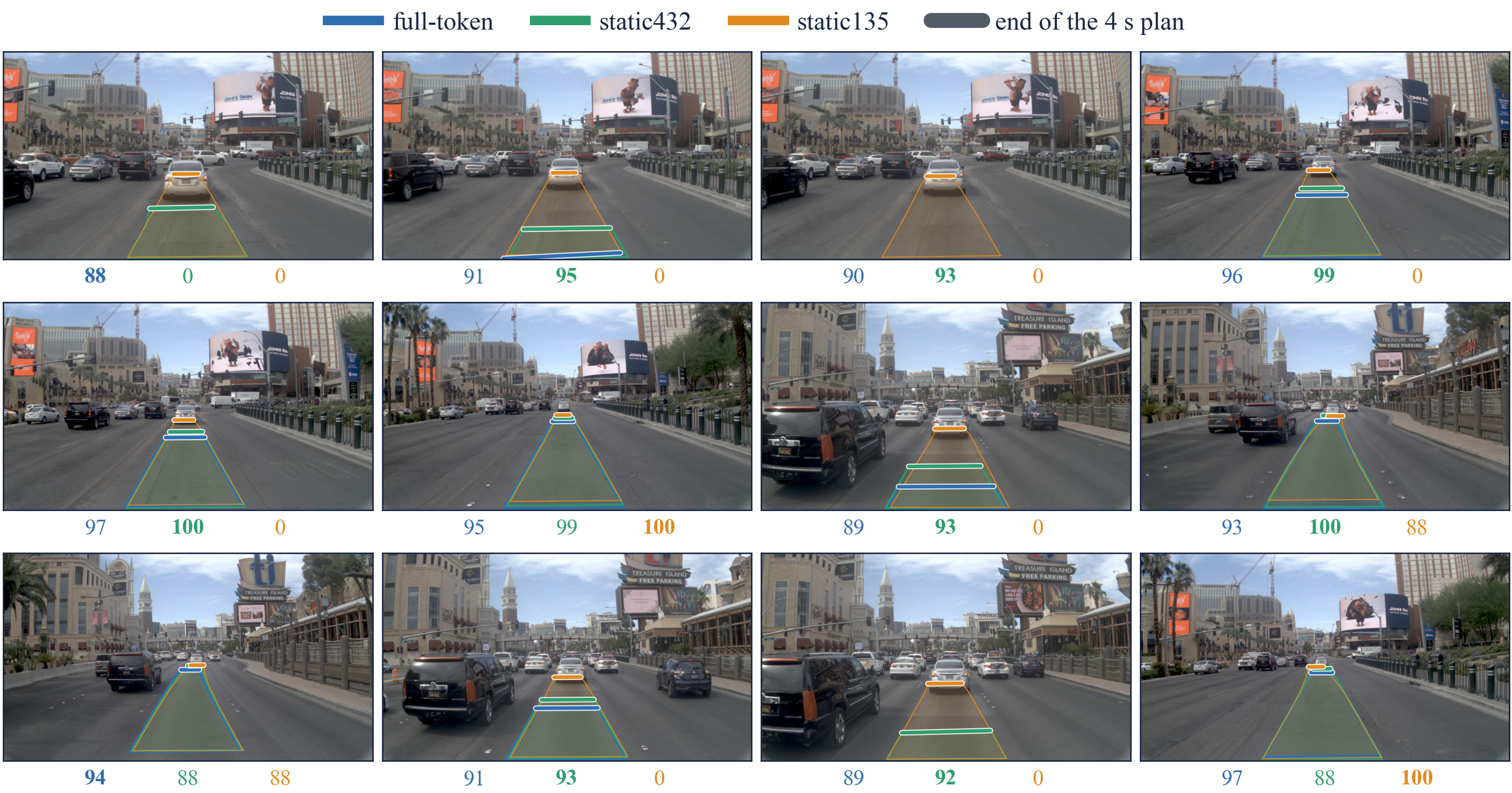}
    \caption{
    \textbf{Budget-dependent trajectories in the camera view.}
    Twelve scenes are selected from the $88$ scenes with synchronized camera data according to the separation between the three four-second trajectory endpoints.
    Semantic inputs are identical across configurations; only $b_t$ changes.
    }
    \label{fig:case_camera}
\end{figure*}

\textbf{Changing only the compute budget can visibly alter the predicted trajectory.}
Fig.~\ref{fig:case_camera} visualizes the same effect directly in the front-camera view.
The retained log contains $88$ scenes with synchronized camera data, from which we select the $12$ scenes with the largest separation between the three four-second trajectory endpoints.
Each panel keeps the semantic input unchanged and varies only the inference configuration $b_t$, matching Eq.~\ref{eq:budgeted_inference}.
The resulting trajectory ribbons therefore visualize the effect of the term after the semicolon in
$f_\theta^{\mathrm{WA}}(\cdot\mid s_t,o_t,g_t;b_t)$.

This visualization makes the conditional direction particularly concrete.
A conventional pipeline chooses a token budget and then obtains an execution latency and trajectory.
SlackDrive reverses the decision order at runtime: realized latency determines the admissible compute set first, after which the highest-utility feasible budget is selected.
For example, when the preceding vehicle brakes abruptly, the controller first preserves completion within the required latency envelope, then uses any remaining headroom to retain additional visual evidence for the next plan.

\begin{figure*}[t]
    \centering
    \includegraphics[width=\textwidth]{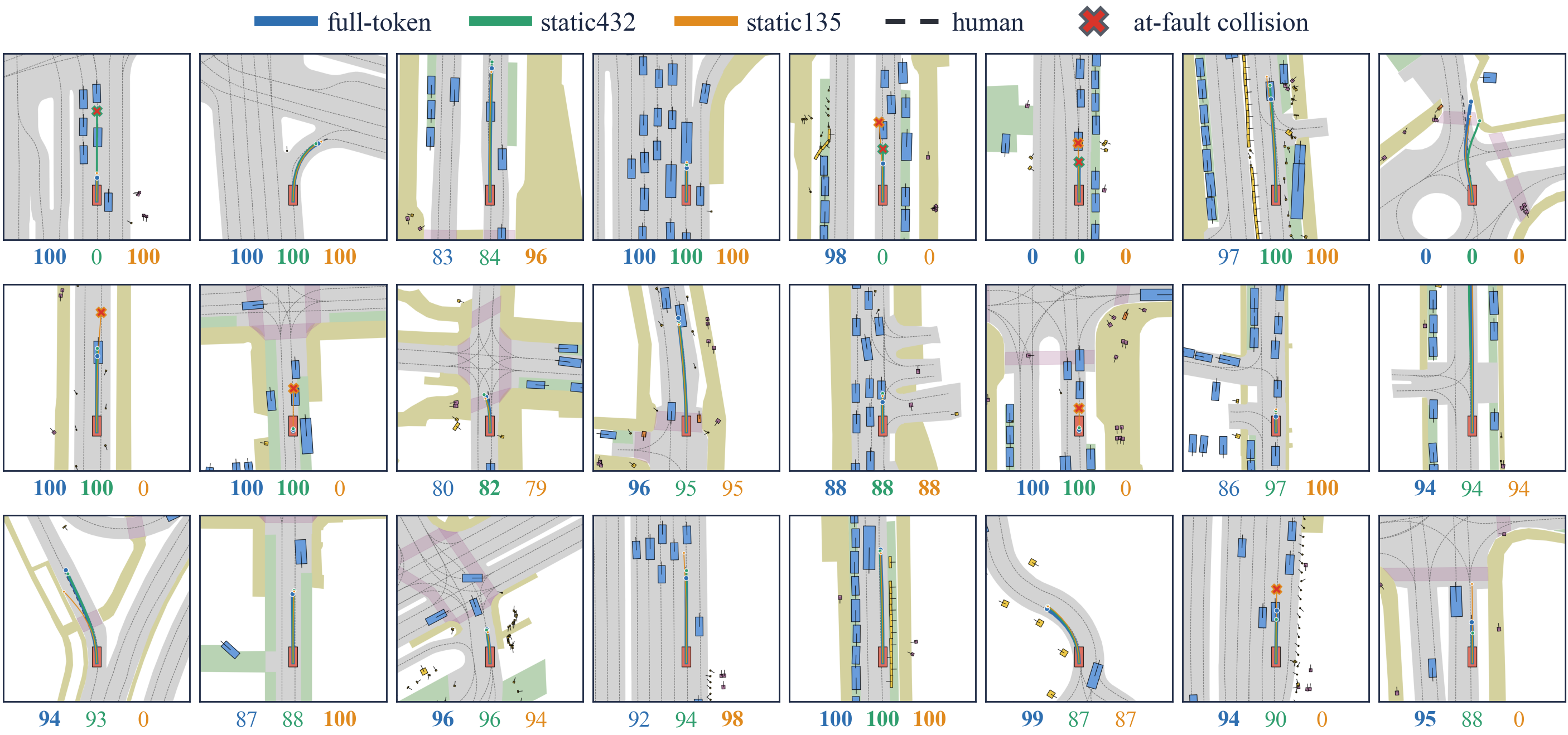}
    \caption{
    \textbf{Static-budget examples.}
    Twenty-four scenes are uniformly sampled from all $10{,}439$ navtest scenes with a fixed random seed and without outcome-based filtering.
    The three trajectories frequently overlap, indicating that moderate budget reduction often preserves the resulting plan.
    }
    \label{fig:case_random}
\end{figure*}

\textbf{Reducing compute usually leaves the resulting plan nearly unchanged.}
Fig.~\ref{fig:case_random} shows a fixed-seed uniform sample of $24$ scenes drawn from all $10{,}439$ scenes without conditioning on the outcome.
In most examples, trajectories from the three operating points overlap closely.
This observation complements Eq.~\ref{eq:robust_feasible}: when increasing runtime load contracts $\mathcal F_{t+1}$, moving to a lower operating point often reduces execution cost without materially changing the planned trajectory.
The population statistics in Fig.~\ref{fig:case_population} provide the corresponding quantitative view.

This redundancy is what makes latency-conditioned compute reduction useful.
On a simple lane-following segment, for instance, retaining every available visual token may change neither the trajectory nor the planning score.
If concurrent workloads reduce the available runtime headroom, SlackDrive can therefore move to a cheaper profiled configuration while preserving the control period instead of executing a previously chosen high-compute configuration and discovering the latency violation afterward.

\begin{figure*}[t]
    \centering
    \includegraphics[width=\textwidth]{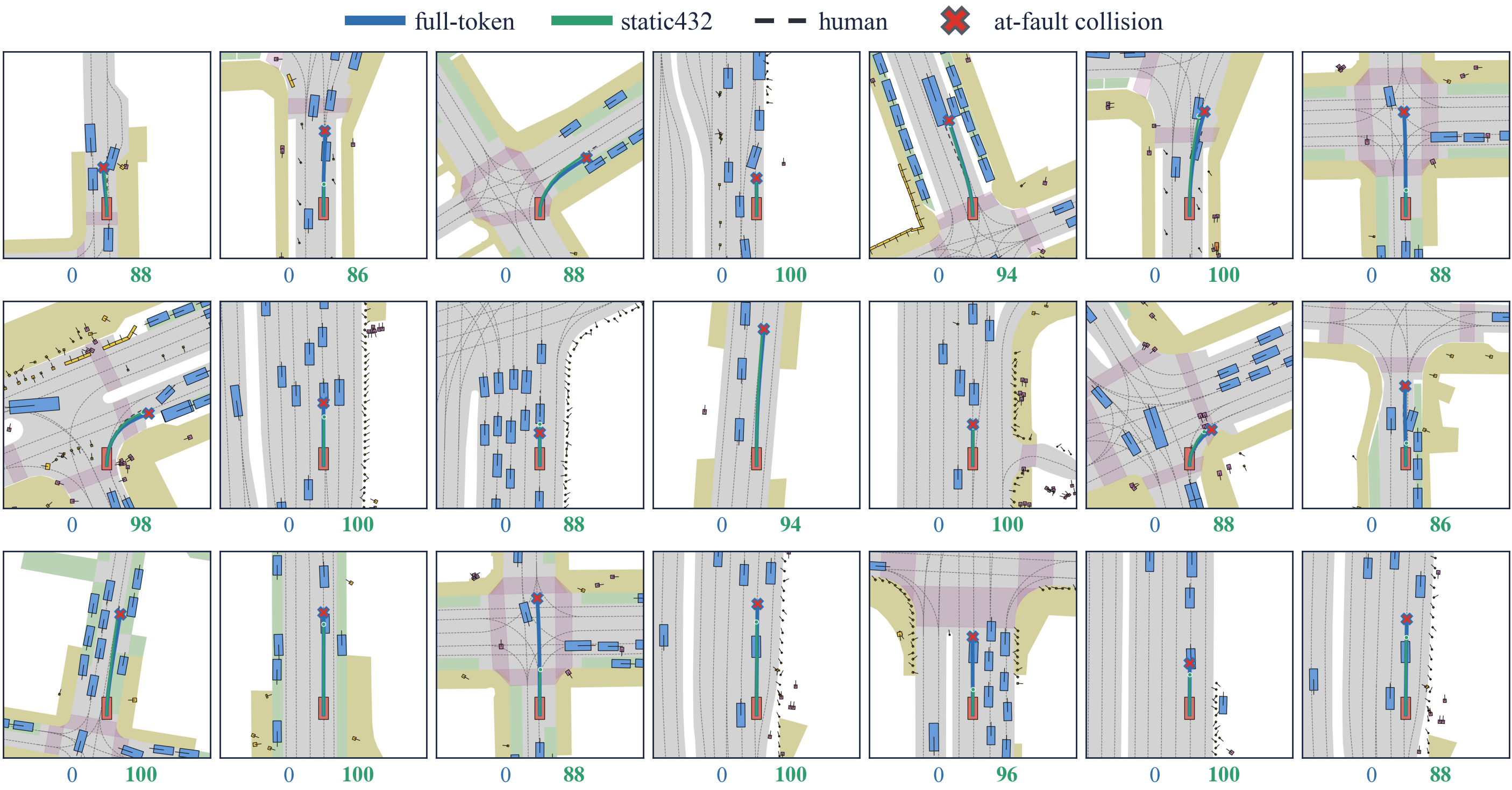}
    \caption{
    \textbf{More compute does not guarantee higher per-scene utility.}
    All $21$ scenes satisfying the selection rule are shown.
    Full-token inference collides, while Static@432 remains collision-free with EPDMS $\ge80$.
    These examples establish existence rather than a population-level safety trend.
    }
    \label{fig:case_more_not_safer}
\end{figure*}

\textbf{More computation is not guaranteed to improve every individual scene.}
Fig.~\ref{fig:case_more_not_safer} contains all $21$ scenes in which full-token inference incurs an at-fault collision while Static@432 remains collision-free with EPDMS at least $80$.
Without the EPDMS threshold, this direction occurs in $26$ of $10{,}439$ scenes.
These examples provide direct evidence for the statement preceding Eq.~\ref{eq:pareto_profile} that additional raw computation should not itself be treated as the optimization objective.
SlackDrive therefore maximizes the profiled utility $U(b)$ within the admissible set instead of maximizing the budget value.

The interpretation here is intentionally limited.
These scenes do not imply that reducing compute generally improves safety.
Instead, they show why Eq.~\ref{eq:chance_objective} optimizes utility rather than computation itself.
In a complex merge or multi-agent interaction, for example, changing the compute configuration may alter the generated trajectory non-monotonically even though the population-level utility favors the larger configuration.

\begin{figure*}[t]
    \centering
    \includegraphics[width=\textwidth]{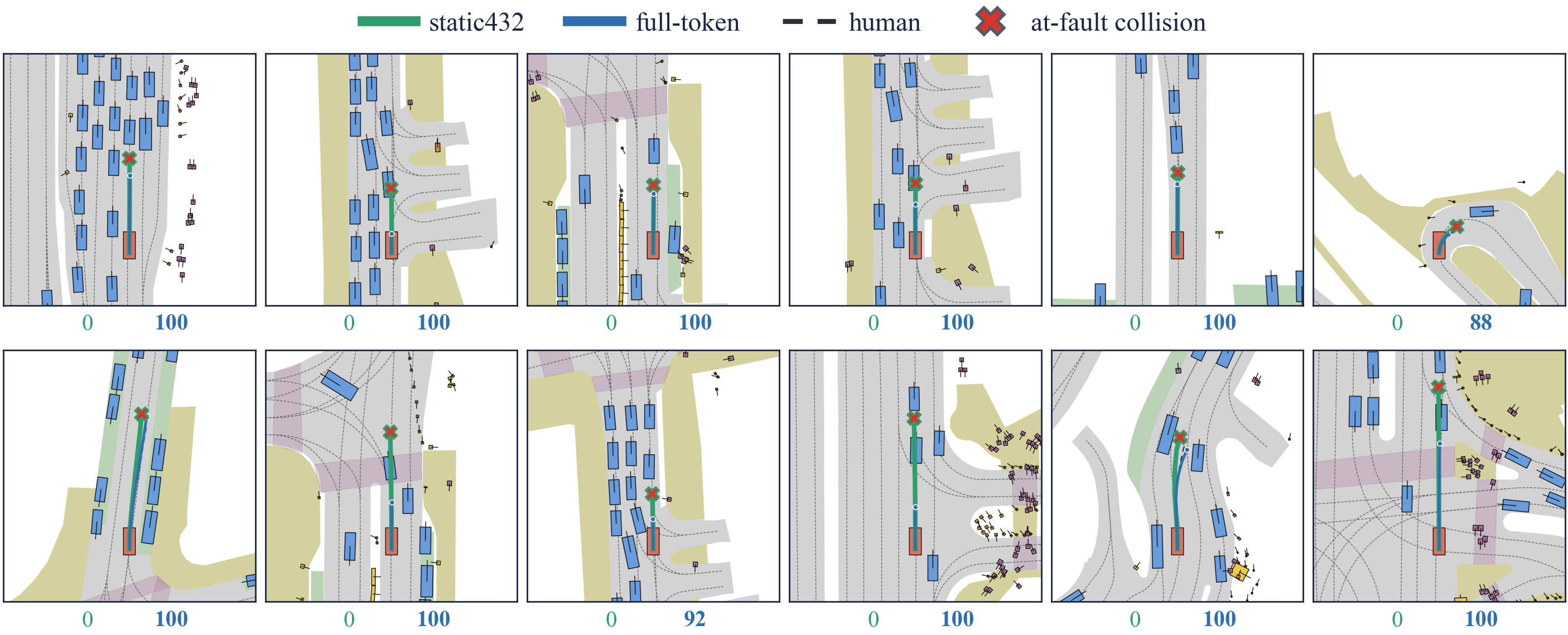}
    \caption{
    \textbf{The population trend favors additional compute.}
    Twelve fixed-seed samples are shown from $315$ scenes in which Static@432 collides while full-token inference remains collision-free with EPDMS $\ge80$.
    }
    \label{fig:case_converse}
\end{figure*}

\textbf{Higher-compute configurations remain safer on average.}
Fig.~\ref{fig:case_converse} shows the converse direction.
There are $315$ scenes in which Static@432 incurs an at-fault collision while full-token inference is collision-free with EPDMS at least $80$; we uniformly sample $12$ using a fixed random seed.
This direction is substantially more common than the converse in Fig.~\ref{fig:case_more_not_safer}, explaining why the population utility satisfies
$U(\text{full})>U(\text{432})>U(\text{135})$.
The Pareto profile therefore retains all three operating points even though isolated scenes may violate this ordering.

These converse cases also explain why SlackDrive reclaims slack instead of simply minimizing computation.
When runtime conditions permit a larger operating point, the additional compute has measurable expected utility.
For example, an occluded road user or a dense interaction at an intersection may benefit from retaining a richer visual representation.
The role of the online policy is therefore to recover such compute whenever the current latency envelope permits it.

\begin{figure*}[t]
    \centering
    \includegraphics[width=\textwidth]{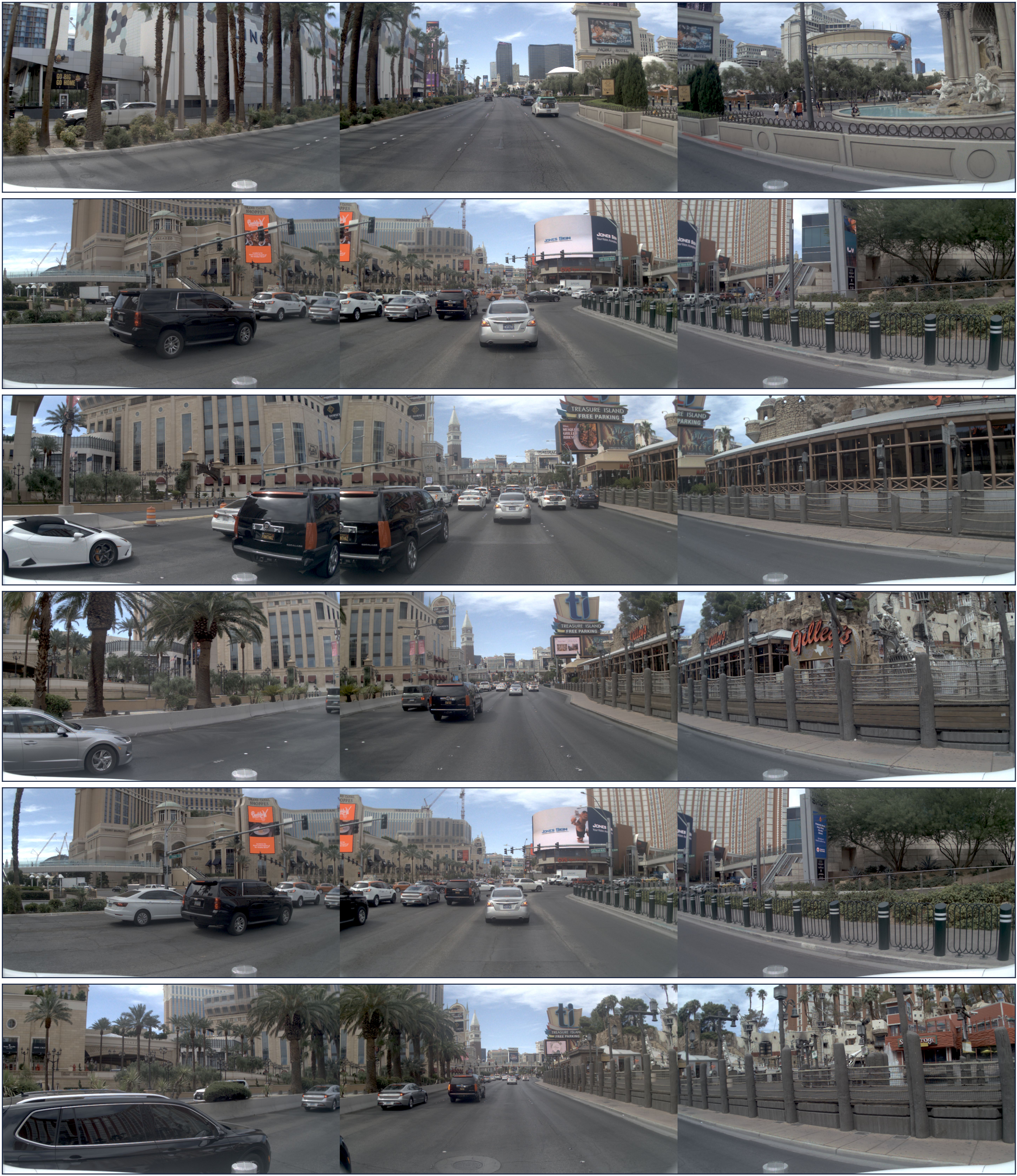}
    \caption{
    \textbf{Representative raw sensor observations.}
    Six observations are evenly spaced in token order from the available synchronized sensor log.
    The same visual input underlies all three compute configurations.
    }
    \label{fig:case_sensor}
    \vspace{-9mm}
\end{figure*}

\textbf{All compute configurations operate on the same underlying sensor stream.}
Related MLLM studies similarly adapt visual computation to task-relevant regions or queries for fine-grained perception~\cite{shi2026catching,shi2026q}.
Fig.~\ref{fig:case_sensor} shows six evenly spaced observations from the $88$ scenes for which raw camera data are available.
These images correspond to the observation $o_t$ in Eq.~\ref{eq:driving_formulation}.
For the visual-token actuator used in the main experiments, Eq.~\ref{eq:actuator_map} instantiates the abstract budget through
$\xi(b)=N_{\mathrm{vis}}(b)$.
The $135$, $432$, and full-token operating points therefore alter how much visual representation is retained from the same observation rather than changing the observation itself.

This distinction is important for interpreting latency-conditioned allocation.
When a cyclist emerges from behind parked vehicles, for example, SlackDrive does not alter the sensor input, navigation context, or semantic driving objective.
Instead, it changes only the amount of profiled computation allocated to the next forward pass according to the current latency envelope and recent runtime state.
The resulting trajectory therefore reflects a different allocation of inference capacity over the same underlying observation, rather than a change in perception input or task definition.
In this sense, SlackDrive adapts how much computation is spent on the current scene while preserving what the model observes and what the driving policy is required to predict.

\subsection{Profile Construction}
\label{sec:appendix_profile}

\textbf{The three operating points form the complete Pareto profile used in the case analysis.}
Table~\ref{tab:profile_appendix} reports the tuple
$\mathcal P=\{(b,L_0(b),U(b))\}$
from Eq.~\ref{eq:pareto_profile}.
Both nominal latency and mean utility increase across the three configurations, so none is dominated and
$\mathcal B_{\mathrm F}=\mathcal B$.
The same $L_0(b)$ values serve as the normalization terms in Eq.~\ref{eq:load_factor} and the nominal latency terms in Eq.~\ref{eq:robust_feasible}.

\begin{table}[t]
    \centering
    \small
    \caption{
    \textbf{Static compute profile used in the case analysis.}
    $L_0(b)$ is nominal latency and $U(b)$ is mean NAVSIM-v2 EPDMS over the analyzed navtest scenes.
    }
    \label{tab:profile_appendix}
    \setlength{\tabcolsep}{7pt}
    \renewcommand{\arraystretch}{1.12}
    \begin{tabular}{lccc}
        \toprule
        \textbf{Budget} & \textbf{Configuration} & $\boldsymbol{L_0(b)}$ \textbf{(ms)} & $\boldsymbol{U(b)}$ \\
        \midrule
        $b_1$ & Static@135 & 54  & 62.88 \\
        $b_2$ & Static@432 & 84  & 81.18 \\
        $b_3$ & Full-token & 226 & 88.48 \\
        \bottomrule
    \end{tabular}
\end{table}

\textbf{The profile captures expected utility, not a per-scene monotonicity assumption.}
The population ordering in Table~\ref{tab:profile_appendix} is sufficient for constructing the frontier even though Fig.~\ref{fig:case_more_not_safer} shows isolated scenes with the opposite ordering.
If a larger profile contains a configuration that is both slower and lower utility on the profiling split, Eq.~\ref{eq:pareto_profile} removes that point before online allocation.

The reported utilities in Table~\ref{tab:profile_appendix} are computed from the navtest scenes used for this appendix analysis.
In the final evaluation protocol, the utility values used to construct $\mathcal P$ should be estimated from a profiling or validation split disjoint from the final test scenes.
For the three operating points analyzed here, the latency and utility ordering is monotonic, so this distinction does not affect the qualitative frontier structure reported above.

\subsection{Collision Outcome Statistics}
\label{sec:appendix_collision}

\textbf{Additional compute is beneficial far more often than harmful, but neither direction is universal.}
Table~\ref{tab:collision_outcomes} reports the complete pairwise collision outcomes across the $10{,}439$ scenes.
For Static@135 versus Static@432, the cheaper configuration alone collides in $1{,}133$ scenes, compared with $74$ scenes in the converse direction.
For Static@432 versus full-token inference, the corresponding counts are $352$ and $26$.
These population statistics provide the base rates behind the selected galleries.

\begin{table}[t]
    \centering
    \scriptsize
    \caption{
    \textbf{Pairwise collision outcomes over all $10{,}439$ scenes.}
    ``Cheap only'' denotes an at-fault collision from the lower-compute configuration but not the higher-compute configuration; ``expensive only'' denotes the converse.
    }
    \label{tab:collision_outcomes}
    \setlength{\tabcolsep}{4.5pt}
    \renewcommand{\arraystretch}{1.12}
    \begin{tabular}{lrrrr}
        \toprule
        \textbf{Budget pair}
        & \textbf{Both safe}
        & \textbf{Cheap only}
        & \textbf{Expensive only}
        & \textbf{Both collide} \\
        \midrule
        Static@432 vs. Full-token
        & 9,888 & 352 & 26 & 149 \\

        Static@135 vs. Static@432
        & 8,702 & 1,133 & 74 & 426 \\

        Static@135 vs. Full-token
        & 8,751 & 1,414 & 29 & 145 \\
        \bottomrule
    \end{tabular}
\end{table}

\textbf{The case galleries expose the utility structure that the online allocator acts upon.}
The $1{,}052$ scenes in Fig.~\ref{fig:case_cheapest} show why selecting the highest-utility feasible configuration differs from always selecting the fastest one.
The random examples in Fig.~\ref{fig:case_random} show why moving to a cheaper configuration is often inexpensive when the feasible set contracts.
The rare converse cases in Fig.~\ref{fig:case_more_not_safer} show why compute itself is not the optimization objective.
Together, these results motivate the two branches of Eq.~\ref{eq:full_policy}: use the highest-utility admissible configuration when the feasible set is nonempty, and fall back to the fastest available configuration only when no profiled operating point satisfies the current latency envelope.

\end{document}